\documentclass{article}

\usepackage[utf8]{inputenc}
\usepackage{main} 
\usepackage{microtype}
\usepackage{times}
\usepackage{latexsym}
\usepackage{algorithm}
\usepackage{algorithmic}
\usepackage{amsmath}
\usepackage{amssymb}
\usepackage{amsfonts}
\usepackage{mathtools}
\usepackage{bm}

\usepackage{graphicx}
\usepackage{subcaption}
\usepackage{float}
\usepackage{wrapfig}

\usepackage{booktabs}
\usepackage{array}
\usepackage{multirow}
\usepackage{multicol}
\usepackage{makecell}
\usepackage{longtable}
\usepackage{tabularx}
\usepackage{colortbl}
\usepackage{arydshln}

\usepackage{color}
\usepackage{xcolor}
\definecolor{mydarkblue}{rgb}{0,0.08,0.45}
\definecolor{wkblue}{rgb}{0.2, 0.3, 0.6}
\definecolor{meta-color}{rgb}{0.5, 0.5, 0.5}
\definecolor{bgblue}{RGB}{245,243,253}
\definecolor{ttblue}{RGB}{91,194,224}
\definecolor{mybrown}{RGB}{128,64,0}
\definecolor{titlecolor}{HTML}{4c9cff}
\definecolor{coolblue3}{rgb}{0.91, 0.94, 0.98}
\definecolor{myblue}{rgb}{0.9, 0.1, 0.94}
\definecolor{mygreen}{rgb}{0.64, 0.56, 0.88}
\definecolor{myyellow}{rgb}{0.68, 0.6, 0.1}
\definecolor{fancygreen}{rgb}{0.33, 0.68, 0.20}
\definecolor{salmon}{rgb}{0.94, 0.52, 0.49}
\definecolor{tablegreen}{rgb}{0.82, 0.94, 0.75}
\definecolor{tableblue}{rgb}{0.81, 0.90, 0.94}
\definecolor{tablered}{rgb}{0.97, 0.85, 0.85}
\definecolor{tableorange}{rgb}{0.96, 0.85, 0.81}

\usepackage{enumitem}
\usepackage{footnote}
\usepackage{lipsum}
\usepackage{setspace}

\usepackage{geometry}
\usepackage{fancyhdr}
\usepackage{lscape}
\usepackage{rotating}

\usepackage{algorithm}
\usepackage{algorithmic}

\usepackage[most]{tcolorbox}
\usepackage[framemethod=tikz]{mdframed}
\usepackage{awesomebox}

\usepackage{natbib}
\usepackage{url}
\usepackage[colorlinks=true,linkcolor=mydarkblue,citecolor=mydarkblue,filecolor=mydarkblue,urlcolor=mydarkblue]{hyperref}

\usepackage{bbding}
\usepackage{imakeidx}
\makeindex
\usepackage[toc]{multitoc}
\usepackage[edges]{forest}
\usepackage[normalem]{ulem}
\usepackage{fontawesome5}
\usepackage{blindtext}
\usepackage{pgfplots}
\usepackage{pgfplotstable}

\usepackage{listings}
\usepackage{listingsutf8}
\newcommand\JSONnumbervaluestyle{\color{blue}}
\newcommand\JSONstringvaluestyle{\color{red}}

\newif\ifcolonfoundonthisline

\makeatletter
\lstdefinestyle{json}
{
  showstringspaces    = false,
  keywords            = {false,true},
  alsoletter          = 0123456789.,
  morestring          = [s]{"}{"},
  stringstyle         = \ifcolonfoundonthisline\JSONstringvaluestyle\fi,
  MoreSelectCharTable =%
    \lst@DefSaveDef{`:}\colon@json{\processColon@json},
  basicstyle          = \ttfamily,
  keywordstyle        = \ttfamily\bfseries,
}

\newcommand\processColon@json{%
  \colon@json%
  \ifnum\lst@mode=\lst@Pmode%
    \global\colonfoundonthislinetrue%
  \fi
}

\lst@AddToHook{Output}{%
  \ifcolonfoundonthisline%
    \ifnum\lst@mode=\lst@Pmode%
      \def\lst@thestyle{\JSONnumbervaluestyle}%
    \fi
  \fi
  \lsthk@DetectKeywords%
}

\lst@AddToHook{EOL}%
  {\global\colonfoundonthislinefalse}
\makeatother

\newtcolorbox{myboxi}[1][]{
  breakable,
  title=#1,
  colback=red!5,
  colbacktitle=red!5,
  coltitle=black,
  fonttitle=\bfseries,
  bottomrule=0pt,
  toprule=0pt,
  leftrule=2pt,
  rightrule=2pt,
  titlerule=0pt,
  arc=0pt,
  outer arc=0pt,
  colframe=red,
}

\newtcolorbox{myboxnote}[1][]{
  breakable,
  title=#1,
  colback=orange!0,
  colbacktitle=orange!0,
  coltitle=black,
  fonttitle=\bfseries,
  bottomrule=0pt,
  toprule=0pt,
  leftrule=2pt,
  rightrule=2pt,
  titlerule=0pt,
  arc=0pt,
  outer arc=0pt,
  colframe=orange,
}

\newtcolorbox{myboxii}[1][]{
  breakable,
  freelance,
  title=#1,
  colback=white,
  colbacktitle=white,
  coltitle=black,
  fonttitle=\bfseries,
  bottomrule=0pt,
  boxrule=0pt,
  colframe=white,
  overlay unbroken and first={
  \draw[red!75!black,line width=3pt]
    ([xshift=5pt]frame.north west) --
    (frame.north west) --
    (frame.south west);
  \draw[red!75!black,line width=3pt]
    ([xshift=-5pt]frame.north east) --
    (frame.north east) --
    (frame.south east);
  },
  overlay unbroken app={
  \draw[red!75!black,line width=3pt,line cap=rect]
    (frame.south west) --
    ([xshift=5pt]frame.south west);
  \draw[red!75!black,line width=3pt,line cap=rect]
    (frame.south east) --
    ([xshift=-5pt]frame.south east);
  },
  overlay middle and last={
  \draw[red!75!black,line width=3pt]
    (frame.north west) --
    (frame.south west);
  \draw[red!75!black,line width=3pt]
    (frame.north east) --
    (frame.south east);
  },
  overlay last app={
  \draw[red!75!black,line width=3pt,line cap=rect]
    (frame.south west) --
    ([xshift=5pt]frame.south west);
  \draw[red!75!black,line width=3pt,line cap=rect]
    (frame.south east) --
    ([xshift=-5pt]frame.south east);
  },
}

\mdfdefinestyle{mystyle}{%
  rightline=true,
  innerleftmargin=10,
  innerrightmargin=10,
  outerlinewidth=3pt,
  topline=false,
  rightline=true,
  bottomline=false,
  skipabove=\topsep,
  skipbelow=\topsep
}

\usepackage{pifont}
\newcommand{\cmark}{\ding{51}}  
\newcommand{\xmark}{\ding{55}}  

\DeclareCaptionFont{black}{\color{black}}

\newenvironment{itemize*}%
 {\leftmargini=10pt\begin{itemize}%
  \setlength{\itemsep}{0pt}%
  \setlength{\parskip}{0pt}%
  }%
 {\end{itemize}}
\newenvironment{enumerate*}%
 {\begin{enumerate}%
  \setlength{\itemsep}{0pt}%
  \setlength{\parskip}{0pt}}%
 {\end{enumerate}}

\usepackage{etoolbox}
\newcounter{bibcount}
\makeatletter
\patchcmd{\@lbibitem}{\item[}{\item[\hfil\stepcounter{bibcount}{[\thebibcount]}}{}{}
\renewcommand\NAT@bibsetup%
  [1]{\setlength{\leftmargin}{\bibhang}\setlength{\itemindent}{-\parindent}%
      \setlength{\itemsep}{\bibsep}\setlength{\parsep}{\z@}}
\makeatother

\definecolor{myblue}{rgb}{0.9, 0.1, 0.94}
\definecolor{mygreen}{rgb}{0.64, 0.56, 0.88}
\definecolor{myyellow}{rgb}{0.68, 0.6, 0.1}
\definecolor{fancygreen}{rgb}{0.33, 0.68, 0.20}
\definecolor{salmon}{rgb}{0.94, 0.52, 0.49}
\definecolor{tablegreen}{rgb}{0.82, 0.94, 0.75}
\definecolor{tableblue}{rgb}{0.81, 0.90, 0.94}
\definecolor{tablered}{rgb}{0.97, 0.85, 0.85}
\definecolor{tableorange}{rgb}{0.96, 0.85, 0.81}

\usepackage{main}

\begin{document}


\newcommand{\modelname}{\textit{AlphaEval}\xspace}

\title{\textit{AlphaEval}: Evaluating Agents in Production}

\author{\mdseries%
Pengrui Lu\textsuperscript{*1,3,4}\quad
Bingyu Xu\textsuperscript{*2,6}\quad
Wenjun Zhang\textsuperscript{*2,5}\quad
Shengjia Hua\textsuperscript{*2}\quad
Xuanjian Gao\textsuperscript{2}\quad
Ranxiang Ge\textsuperscript{2}\\
Lyumanshan Ye\textsuperscript{3,4}\quad
Linxuan Wu\textsuperscript{2}\quad
Yiran Li\textsuperscript{2}\quad
Junfei Fish Yu\textsuperscript{9}\quad
Yibo Zhang\textsuperscript{11}\quad
Ruixin Li\textsuperscript{11} \quad 
Manxiang Li\textsuperscript{11}\\
Xiao Han\textsuperscript{11}\quad
Xiaocong Zhou\textsuperscript{12}\quad
Guangyao Chi\textsuperscript{7}\quad
Zisheng Chen\textsuperscript{8}\quad
Kaishen Chen\textsuperscript{10} \quad
Kun Wang\textsuperscript{10}\quad
Qihua Xu\textsuperscript{10} \\
Fengyue Meng\textsuperscript{2}\quad
Yuchen Ni\textsuperscript{2}\quad
Jiajun Li\textsuperscript{2}\quad
Jinxiu Liu\textsuperscript{2} \quad
Danfeng Zhang\textsuperscript{2}\quad
Jingru Zhao\textsuperscript{2}\quad
Pengfei Liu\textsuperscript{$\dagger$1,3,4}\vspace{4pt}\\
{\footnotesize \textsuperscript{1}SII\enspace \textsuperscript{2}MiraclePlus\enspace \textsuperscript{3}SJTU\enspace \textsuperscript{4}GAIR\enspace \textsuperscript{5}HIT\enspace \textsuperscript{6}UCAS\enspace \textsuperscript{7}LangCore\enspace \textsuperscript{8}Jiqizhixin\enspace \textsuperscript{9}HunterAI\enspace \textsuperscript{10}CinoCore\enspace \textsuperscript{11}KuaFuAI\enspace \textsuperscript{12}POET}}

\date{}

\maketitle


\pagestyle{fancy}
\thispagestyle{fancy}
\fancyhead{}
\lhead{
  \raisebox{-0.3cm}{\includegraphics[height=0.95cm]{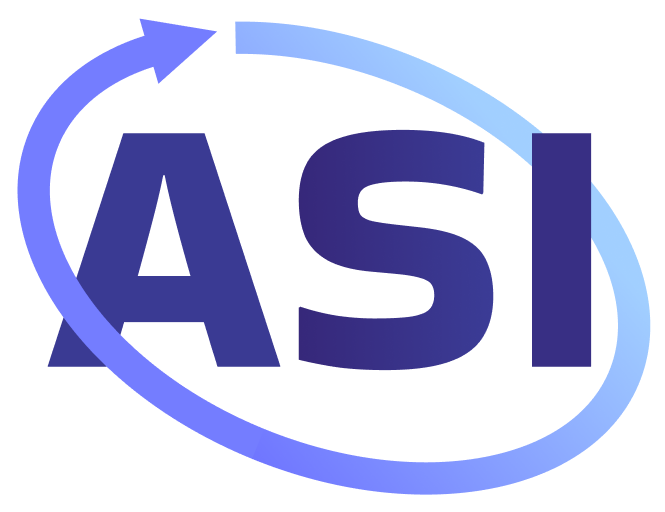}}
}
\rhead{%
  \raisebox{-0.2cm}{\includegraphics[height=0.7cm]{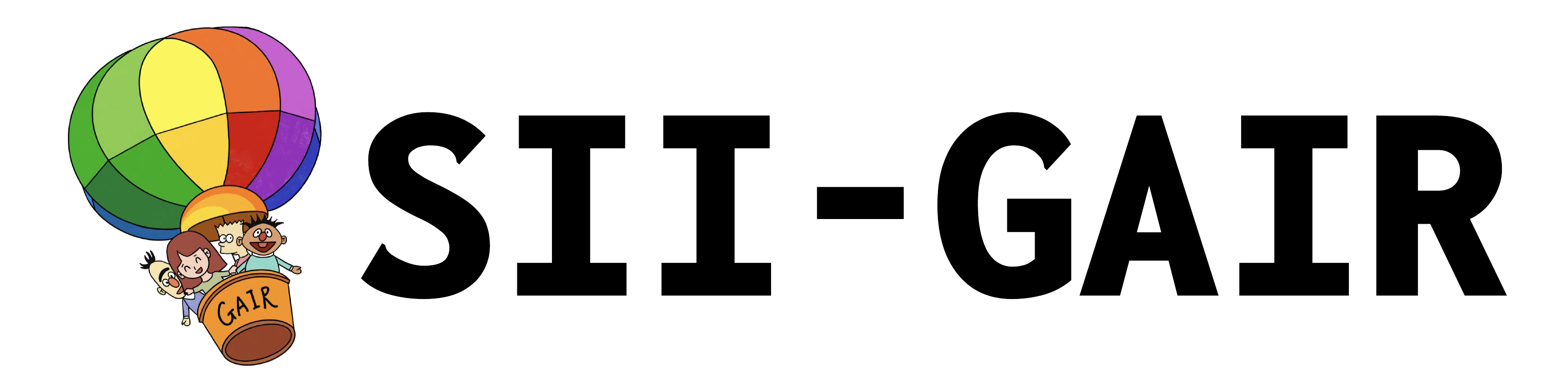}}%
}
\renewcommand{\headrulewidth}{0pt}
\setlength{\headsep}{2mm}

\renewcommand{\thefootnote}{}
\footnotetext{* Equal contribution.}
\footnotetext{† Corresponding author.}
\vspace{-20pt}


\begin{center}
    \href{https://github.com/GAIR-NLP/AlphaEval}{\textcolor{black}{\faGithub\ AlphaEval}}
    \quad
    \href{https://alphaeval.ai}{\raisebox{-.15em}{\includegraphics[height=1em]{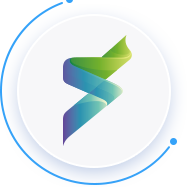}}\ Homepage}
\end{center}

\vspace{20pt}


\begin{abstract}
The rapid deployment of AI agents in commercial settings has outpaced the development of evaluation methodologies that reflect production realities. Existing benchmarks measure agent capabilities through retrospectively curated tasks with well-specified requirements and deterministic metrics---conditions that diverge fundamentally from production environments where requirements contain implicit constraints, inputs are heterogeneous multi-modal documents with information fragmented across sources, tasks demand undeclared domain expertise, outputs are long-horizon professional deliverables, and success is judged by domain experts whose standards evolve over time.
We present \modelname, a production-grounded benchmark of \textbf{94} tasks sourced from \textbf{seven} companies deploying AI agents in their core business, spanning six O*NET (\href{https://www.onetonline.org/}{Occupational Information Network}) domains.
Unlike model-centric benchmarks, \modelname evaluates \textit{complete agent products}---Claude Code, Codex, etc.---as commercial systems, capturing performance variations invisible to model-level evaluation.
Our evaluation framework covers multiple paradigms (LLM-as-a-Judge, reference-driven metrics, formal verification, rubric-based assessment, automated UI testing, etc.), with individual domains composing multiple paradigms.
Beyond the benchmark itself, we contribute a \textit{requirement-to-benchmark construction framework}---a systematic methodology that transforms authentic production requirements into executable evaluation tasks in minimal time. This framework standardizes the entire pipeline from requirement to evaluation, providing a reproducible, modular process that any organization can adopt to construct production-grounded benchmarks for their own domains.

\vspace{-0.2em}
\begin{figure}[htbp]
    \centering
\includegraphics[width=0.8\textwidth]{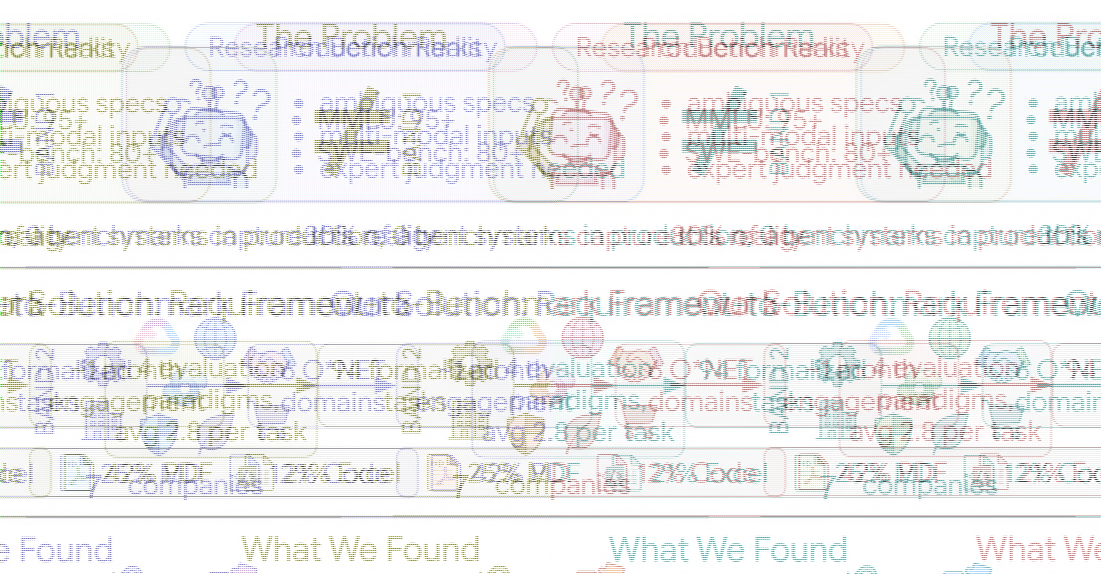}
    \caption{Overview of \modelname. \textbf{Band~1}: The gap between research benchmarks and production reality. \textbf{Band~2}: Our requirement-to-benchmark framework transforms production requirements into 94 formalized tasks across 6 O*NET domains. \textbf{Band~3}: Key findings---best agent scores 64.41/100, scaffold matters as much as model, and six production-specific failure modes.}
    \label{fig:overview}
\end{figure}

\end{abstract}




\clearpage  

\pagestyle{fancy}
\lhead{\rightmark}
\renewcommand{\headrulewidth}{0.7pt}
\setlength{\headsep}{5mm}

\section{Introduction}
\label{sec:introduction}

The hardest part of deploying an AI agent is not building the agent---it is knowing whether the agent works. As organizations move from prototypes to production, constructing reliable evaluation infrastructure has emerged as a dominant engineering challenge~\citep{pan2025measuring}. Yet the research community has devoted remarkably little attention to \textit{evaluating agents as they are actually deployed in production}.
Existing benchmarks like SWE-bench~\citep{jimenez2024swebench}, WebArena~\citep{zhou2024webarena}, and OSWorld~\citep{xie2024osworld} have been instrumental in driving agent capabilities forward. However, they share a common design philosophy: tasks are curated retrospectively---drawn from already-completed work artifacts such as resolved GitHub issues or archived web sessions---with well-specified requirements and deterministic metrics, diverging fundamentally from production evaluation. A recent systematic analysis of 43 benchmarks covering 72,342 tasks confirms this mismatch quantitatively, revealing that agent development remains heavily programming-centric and misaligned with the occupational categories where human labor and economic value concentrate~\citep{wang2026howwell}.

Production evaluation operates under different constraints: requirements contain implicit constraints and arrive as loosely worded business descriptions; inputs are heterogeneous multi-modal documents with information fragmented across sources; tasks demand undeclared domain expertise; and success is judged by domain experts' subjective assessments. Among practitioners who reported their deployment stage, over 80\% indicated their agent systems are in production or pilot phases~\citep{pan2025measuring}, yet no existing benchmark fully captures this reality. Our survey of 27 AI product companies (Appendix~\ref{sec:appendix_survey}) quantifies this gap: \textbf{63\%} of companies report low confidence in whether model updates actually improve their products, 25.9\% have \textit{no explicit evaluation criteria}, and 70.4\% rely on developers performing testing as a side task. The disconnect is structural along three axes: (1) \textit{Task Under-specification}: benchmarks have explicit goals; production tasks emerge from evolving business needs with implicit constraints invisible to outsiders. (2) \textit{Judgment Subjectivity}: benchmarks evaluate with single-dimension, predefined metrics; production demands multi-faceted quality criteria defined by domain experts. (3) \textit{Continuous Evolution}: benchmarks are built once; production evaluations must continuously evolve.

We address this by partnering with companies deploying AI agents in their core business and AI-focused organizations to capture authentic evaluation scenarios. The result is \modelname,\footnote{Evaluation framework available at \url{https://github.com/GAIR-NLP/AlphaEval}.} a benchmark of \textbf{94} production-grounded tasks together with a reusable \textit{requirement-to-benchmark construction framework} that standardizes the pipeline from authentic production requirements to fully automated evaluations. We evaluate six frontier models---Claude Opus 4.6, GPT-5.2, Gemini 3 Pro Preview, Kimi K2.5, GLM-5, and MiniMax M2.5---deployed through four commercial agent products: Claude Code, Codex, GitHub Copilot, and Cursor, yielding 14 model--scaffold configurations. Our key findings: (1) the best configuration (Claude Code + Opus 4.6) achieves only 64.41/100, revealing a substantial research-production gap; (2) scaffold choice matters as much as model choice---the same Opus 4.6 model scores 64.41 via Claude Code but only 53.45 via Codex, an 11-point spread; and (3) domain performance varies dramatically, from 62.0 (Technology Research) to 30.0 (Human Resources), indicating that no single aggregate score captures production readiness (Figure~\ref{fig:overview}). Our contributions are fourfold:
\begin{itemize}[leftmargin=*,itemsep=2pt,topsep=2pt]
\item \textbf{A requirement-to-benchmark construction framework.} A standardized methodology that transforms authentic production requirements from companies into automated, reproducible evaluation tasks through a modular four-stage pipeline (partner engagement, requirement elicitation, task formalization, iterative validation), enabling systematic and repeatable benchmark construction across diverse professional domains.
\item \textbf{A production-grounded agent benchmark.} 94 tasks from seven companies classified into six O*NET occupational domains, preserving production complexity---ambiguous specifications, multi-modal inputs, long-horizon deliverables, and stakeholder-aligned evaluation criteria that evolve with business needs.
\item \textbf{A unified evaluation framework.} Multiple evaluation paradigms through a standardized API with Docker-based sandboxed execution. Individual domains \textit{compose} multiple paradigms, reflecting the multi-dimensional nature of production quality.
\item \textbf{An empirical analysis of the research-production gap.} Frontier agents exhibit significant degradation on production tasks, with the best configuration achieving only 64.41 on average. We conduct a systematic error analysis of failure cases and provide economic value grounding that translates benchmark scores into dollar-value estimates of professional labor delivered.
\end{itemize}

\section{Preliminaries}
\label{sec:background}

\begin{figure}[t]
    \centering
    \includegraphics[width=0.9\linewidth]{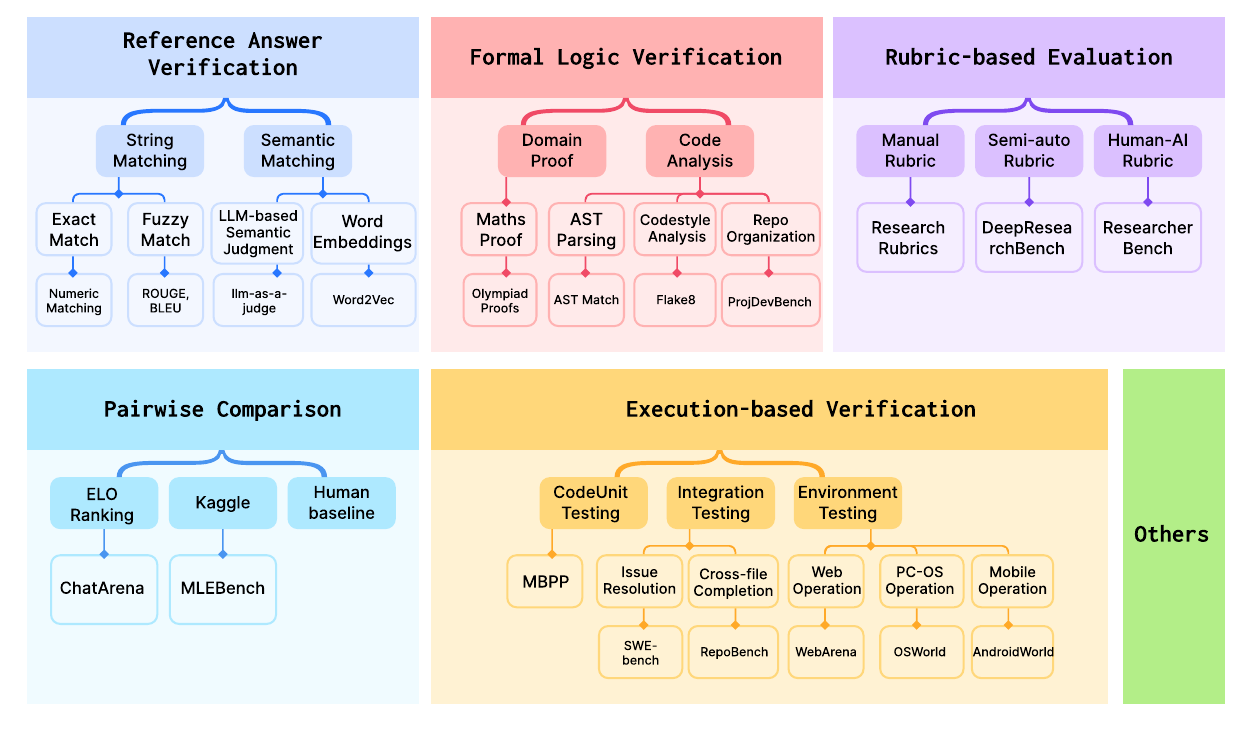}
    \caption{Taxonomy of evaluation methodologies for AI agents. \modelname covers multiple evaluation paradigms (excluding Pairwise Comparison and Others).}
    \label{fig:eval_taxonomy}
\end{figure}

\paragraph{Evaluation Methodology Taxonomy.}
Through systematic analysis of 90+ existing agent-relevant benchmarks, we identify a taxonomy of evaluation methodologies (Figure~\ref{fig:eval_taxonomy}). These span four major paradigms: \textbf{Reference Answer Verification} (string and semantic matching against gold answers), \textbf{Formal Logic Verification} (mathematical proofs and code analysis), \textbf{Rubric-based Evaluation} (manual, semi-automated, or human-AI co-designed scoring rubrics), and \textbf{Execution-based Verification} (code unit testing and environment state testing). Beyond these paradigms, LLM-as-a-Judge~\citep{zheng2023judging} and Agent-as-a-Judge~\citep{zhuge2024agentjudge} serve as cross-cutting evaluation methods that operate \textit{across} paradigms---not merely as semantic matching within Reference Answer Verification, but as flexible judges capable of rubric-based scoring, factual verification, and holistic quality assessment. Notably, most existing benchmarks employ only a single paradigm per task, whereas production evaluation typically requires composing multiple paradigms to capture multi-dimensional quality.

\paragraph{Revisit Existing Agent Benchmarks}
Task-based agent benchmarks~\citep{jimenez2024swebench,zhou2024webarena,xie2024osworld,agencybench2026} enable rigorous capability measurement but curate tasks retrospectively with automatable metrics. Code benchmark surveys~\citep{openbench2024} and protocol-based tool-use benchmarks~\citep{mcpuniverse2025} pursue evaluation rigor but remain anchored in research contexts. Domain-specific evaluation for finance~\citep{fincdm2025}, e-commerce~\citep{ecombench2025}, and healthcare~\citep{csedb2025} remain researcher-curated with predetermined scopes. The rise of LLM-as-a-Judge enables scalable evaluation but introduces reliability challenges~\citep{panickssery2024,zheng2023judging} exacerbated in production settings. Notably, production evaluation spans diverse paradigms---LLM-as-a-Judge, reference-driven metrics, formal verification, rubric-based assessment, and automated UI testing---with most domains combining multiple paradigms, a diversity that existing benchmarks do not capture. Concurrently, \citet{wang2026howwell} systematically map 43 agent benchmarks to all 1,016 U.S.\ occupations via O*NET, finding substantial coverage gaps and proposing coverage, realism, and granular evaluation as design principles---criteria that \modelname operationalizes through production-sourced tasks across six O*NET domains and multi-paradigm evaluation. \modelname addresses this gap by systematizing production evaluation from companies deploying AI agents, preserving production complexity and contributing both a benchmark and a requirement-to-benchmark construction framework. A comprehensive comparison with 90+ benchmarks across seven dimensions is provided in Table~\ref{tab:benchmark_comparison}.

\begingroup
\scriptsize
\setlength{\LTcapwidth}{\linewidth}
\setlength{\tabcolsep}{2.2pt}
\renewcommand{\arraystretch}{1.05}

\begin{longtable}{
  l   
  l   
  l   
  c   
  c   
  c   
  c   
  c   
  c   
  c   
}

\caption{Comprehensive comparison of \modelname with existing benchmarks across seven dimensions that characterize production-level evaluation: \textbf{Production}~=~tasks sourced from real commercial deployments with paying customers; \textbf{Multi-Modal}~=~requires processing multi-modal heterogeneous inputs (PDFs, spreadsheets, images); \textbf{Underspec.}~=~tasks deliberately preserve production-level requirement ambiguity; \textbf{Diverse Eval}~=~employs three or more distinct evaluation paradigms; \textbf{Expert Val.}~=~evaluation criteria co-developed and validated with domain experts; \textbf{Dynamic}~=~benchmark designed for continuous evolution rather than static one-time release; \textbf{Cross-Domain}~=~spans three or more distinct professional domains.}
\label{tab:benchmark_comparison} \\

\toprule
\textbf{Benchmark} &
\textbf{Domain} &
\textbf{Eval Type} &
\rotatebox{70}{\textbf{Production}} &
\rotatebox{70}{\textbf{Multi-Modal}} &
\rotatebox{70}{\textbf{Underspec.}} &
\rotatebox{70}{\textbf{Diverse Eval}} &
\rotatebox{70}{\textbf{Expert Val.}} &
\rotatebox{70}{\textbf{Dynamic}} &
\rotatebox{70}{\textbf{Cross-Domain}} \\
\midrule
\endfirsthead

\multicolumn{10}{l}{\footnotesize\itshape Table~\ref{tab:benchmark_comparison} continued from previous page} \\[3pt]
\toprule
\textbf{Benchmark} &
\textbf{Domain} &
\textbf{Eval Type} &
\rotatebox{70}{\textbf{Production}} &
\rotatebox{70}{\textbf{Multi-Modal}} &
\rotatebox{70}{\textbf{Underspec.}} &
\rotatebox{70}{\textbf{Diverse Eval}} &
\rotatebox{70}{\textbf{Expert Val.}} &
\rotatebox{70}{\textbf{Dynamic}} &
\rotatebox{70}{\textbf{Cross-Domain}} \\
\midrule
\endhead

\midrule
\multicolumn{10}{r}{\footnotesize\itshape Continued on next page} \\
\endfoot

\bottomrule
\endlastfoot

\multicolumn{10}{l}{\cellcolor{coolblue3}\textbf{Software Engineering \& Coding}} \\
\midrule
SWE-bench~\citep{jimenez2024swebench}          & Code         & Rule   & \xmark & \xmark & \xmark & \xmark & \xmark & \xmark & \xmark \\
SWE-bench Multimodal~\citep{swebenchmm2024}    & Code         & Rule   & \xmark & \cmark & \xmark & \xmark & \xmark & \xmark & \xmark \\
Multi-SWE-bench~\citep{multisweb2025}          & Code         & Rule   & \xmark & \xmark & \xmark & \xmark & \xmark & \cmark & \xmark \\
SWE-Lancer~\citep{swelancer2025}               & Code         & Rule   & \cmark & \xmark & \xmark & \xmark & \xmark & \xmark & \xmark \\
SWT-Bench~\citep{swtbench2024}                 & Code         & Rule   & \xmark & \xmark & \xmark & \xmark & \xmark & \xmark & \xmark \\
Terminal-bench~\citep{terminalbench2025}        & Code         & Rule   & \xmark & \xmark & \xmark & \xmark & \xmark & \cmark & \xmark \\
FeatBench~\citep{featbench2025}                & Code         & Rule   & \xmark & \xmark & \xmark & \xmark & \xmark & \cmark & \xmark \\
DevBench~\citep{devbench2024}                  & Code         & R+M+H  & \xmark & \xmark & \xmark & \cmark & \xmark & \xmark & \xmark \\
LongCLI-Bench~\citep{longclibench2026}         & Code         & Rule   & \xmark & \xmark & \xmark & \xmark & \xmark & \xmark & \xmark \\
ProjDevBench~\citep{projdevbench2026}          & Code         & Rule+Model & \xmark & \xmark & \xmark & \xmark & \xmark & \xmark & \xmark \\
\midrule

\multicolumn{10}{l}{\cellcolor{coolblue3}\textbf{Data Science \& ML Engineering}} \\
\midrule
DSBench~\citep{dsbench2024}                    & Code         & Model  & \xmark & \cmark & \xmark & \xmark & \xmark & \xmark & \xmark \\
MLE-bench~\citep{mlebench2024}                 & Code         & Rule   & \xmark & \xmark & \xmark & \xmark & \xmark & \xmark & \xmark \\
KernelBench~\citep{kernelbench2025}            & Code         & Rule   & \xmark & \xmark & \xmark & \xmark & \xmark & \xmark & \xmark \\
DAComp~\citep{dacomp2025}                      & Code+Research  & Rule+Model  & \xmark & \xmark & \xmark & \xmark & \xmark & \xmark & \xmark \\
\midrule

\multicolumn{10}{l}{\cellcolor{coolblue3}\textbf{Code Competition \& Security}} \\
\midrule
LiveCodeBench~\citep{livecodebench2024}        & Code         & Rule   & \xmark & \xmark & \xmark & \cmark & \xmark & \cmark & \xmark \\
CodeElo~\citep{codeforceselo2025}              & Code         & Rule   & \xmark & \xmark & \xmark & \xmark & \xmark & \xmark & \xmark \\
Aider Polyglot~\citep{aiderpolyglot2024}       & Code         & Rule   & \xmark & \xmark & \xmark & \xmark & \xmark & \xmark & \xmark \\
CyBench~\citep{cybench2024}                    & Code         & Rule   & \xmark & \xmark & \xmark & \xmark & \xmark & \xmark & \xmark \\
BountyBench~\citep{bountybench2025}            & Code         & Rule   & \xmark & \xmark & \xmark & \xmark & \xmark & \xmark & \xmark \\
VimGolf-Gym~\citep{vimgolfgym2025}             & Code         & Rule   & \xmark & \xmark & \xmark & \xmark & \xmark & \xmark & \xmark \\
DPAI Arena~\citep{dpaiarena2025}               & Code         & Rule   & \xmark & \xmark & \xmark & \xmark & \xmark & \cmark & \xmark \\
Spring AI Bench~\citep{springaibench2025}       & Code         & Rule   & \xmark & \xmark & \xmark & \xmark & \xmark & \xmark & \xmark \\
AGENTS.md Eval~\citep{agentsmd2026}            & Code         & Rule   & \xmark & \xmark & \xmark & \xmark & \xmark & \xmark & \xmark \\
\midrule

\multicolumn{10}{l}{\cellcolor{coolblue3}\textbf{Tool Use \& Web Interaction}} \\
\midrule
WebArena~\citep{zhou2024webarena}              & Web          & Rule+Model & \xmark & \xmark & \xmark & \xmark & \xmark & \xmark & \cmark \\
AgentBench~\citep{liu2023agentbench}           & Tool         & Rule   & \xmark & \xmark & \xmark & \xmark & \xmark & \xmark & \cmark \\
AgentBoard~\citep{agentboard2024}              & Tool         & Rule   & \xmark & \xmark & \xmark & \xmark & \xmark & \xmark & \cmark \\
$\tau$-bench~\citep{taubench2024}              & Tool         & Rule   & \xmark & \xmark & \xmark & \xmark & \xmark & \xmark & \xmark \\
$\tau^2$-Bench~\citep{tau2bench2025}           & Tool         & Rule   & \xmark & \xmark & \xmark & \xmark & \xmark & \xmark & \xmark \\
TheAgentCompany~\citep{theagentcompany2024}    & Tool         & Rule+Model & \xmark & \cmark & \xmark & \xmark & \xmark & \xmark & \xmark \\
Tool Decathlon~\citep{tooldecathlon2025}       & Tool         & Rule   & \xmark & \xmark & \xmark & \xmark & \xmark & \xmark & \cmark \\
ACEBench~\citep{acebench2025}                  & Tool         & Rule+Model & \xmark & \xmark & \xmark & \cmark & \xmark & \xmark & \cmark \\
MCP-Universe~\citep{mcpuniverse2025}           & Tool         & Rule   & \xmark & \xmark & \xmark & \xmark & \xmark & \xmark & \cmark \\
BFCL~\citep{bfcl2025}                         & Tool         & Rule   & \xmark & \xmark & \xmark & \xmark & \xmark & \cmark & \cmark \\
Context-Bench~\citep{contextbench2025}         & Tool         & Rule   & \xmark & \xmark & \xmark & \xmark & \xmark & \cmark & \xmark \\
Letta Evals~\citep{lettaevals2025}             & Tool         & Rule+Model & \xmark & \xmark & \xmark & \xmark & \xmark & \cmark & \xmark \\
EcomBench~\citep{ecombench2025}                & Tool         & Model  & \xmark & \xmark & \xmark & \xmark & \cmark & \cmark & \xmark \\
DeliveryBench~\citep{deliverybench2025}        & Tool         & Rule   & \xmark & \cmark & \xmark & \xmark & \xmark & \xmark & \xmark \\
WorFBench~\citep{worfbench2024}                & Tool         & Rule   & \xmark & \xmark & \xmark & \xmark & \xmark & \xmark & \cmark \\
BrowseComp~\citep{browsecomp2025}              & Search       & Model  & \xmark & \xmark & \xmark & \xmark & \xmark & \xmark & \cmark \\
AgencyBench~\citep{agencybench2026}            & Code+Tool    & Rule+Model & \xmark & \xmark & \xmark & \xmark & \cmark & \xmark & \cmark \\
HammerBench~\citep{hammerbench2024}            & Tool         & Rule   & \xmark & \xmark & \xmark & \xmark & \xmark & \xmark & \xmark \\
\midrule

\multicolumn{10}{l}{\cellcolor{coolblue3}\textbf{Operating System \& GUI}} \\
\midrule
GAIA~\citep{gaia2023}                         & OS           & Rule   & \xmark & \cmark & \xmark & \xmark & \xmark & \xmark & \cmark \\
OSWorld~\citep{xie2024osworld}                 & OS           & Rule   & \xmark & \cmark & \xmark & \xmark & \xmark & \xmark & \xmark \\
AppWorld~\citep{appworld2024}                  & OS           & Rule   & \xmark & \xmark & \xmark & \xmark & \xmark & \xmark & \xmark \\
WebSuite~\citep{websuite2024}                  & OS           & Rule   & \xmark & \xmark & \xmark & \xmark & \xmark & \xmark & \xmark \\
OSUniverse~\citep{osuniverse2025}              & OS           & R+M+H  & \xmark & \cmark & \xmark & \cmark & \xmark & \xmark & \xmark \\
OdysseyBench~\citep{odysseybench2025}          & OS           & Rule   & \xmark & \xmark & \xmark & \xmark & \xmark & \xmark & \xmark \\
OfficeQA~\citep{officeqa2025}                  & OS           & Rule   & \xmark & \cmark & \xmark & \xmark & \xmark & \xmark & \xmark \\
\midrule

\multicolumn{10}{l}{\cellcolor{coolblue3}\textbf{Scientific Research}} \\
\midrule
EXP-Bench~\citep{expbench2025}                & Research     & Model  & \xmark & \xmark & \xmark & \xmark & \xmark & \xmark & \xmark \\
PaperBench~\citep{paperbench2025}              & Research     & Model  & \xmark & \xmark & \xmark & \xmark & \cmark & \xmark & \xmark \\
CORE-Bench~\citep{corebench2024}               & Research     & Rule   & \xmark & \cmark & \xmark & \xmark & \xmark & \xmark & \cmark \\
Auto-Bench~\citep{autobench2025}               & Research     & Rule   & \xmark & \xmark & \xmark & \xmark & \xmark & \xmark & \xmark \\
ResearchCodeBench~\citep{researchcodebench2025} & Research    & Rule   & \xmark & \xmark & \xmark & \xmark & \xmark & \cmark & \xmark \\
AstaBench~\citep{astabench2025}                & Research     & Rule+Model & \xmark & \xmark & \xmark & \xmark & \xmark & \xmark & \cmark \\
AInsteinBench~\citep{ainsteinbench2025}        & Code+Research & Rule  & \xmark & \xmark & \xmark & \xmark & \xmark & \xmark & \cmark \\
ResearchGym~\citep{researchgym2026}            & Research+Code & Rule  & \xmark & \xmark & \xmark & \xmark & \xmark & \xmark & \xmark \\
\midrule

\multicolumn{10}{l}{\cellcolor{coolblue3}\textbf{Mathematics \& Knowledge}} \\
\midrule
MMLU~\citep{mmlu2020}                         & Knowledge    & Rule   & \xmark & \xmark & \xmark & \xmark & \xmark & \xmark & \cmark \\
GPQA Diamond~\citep{gpqa2023}                  & Knowledge    & Rule   & \xmark & \xmark & \xmark & \xmark & \cmark & \xmark & \cmark \\
MMMU~\citep{mmmu2023}                         & Knowledge    & Rule   & \xmark & \cmark & \xmark & \xmark & \xmark & \xmark & \cmark \\
MathVista~\citep{mathvista2024}                & Math         & Rule   & \xmark & \cmark & \xmark & \xmark & \xmark & \xmark & \xmark \\
FrontierMath~\citep{frontiermath2024}          & Math         & Rule   & \xmark & \xmark & \xmark & \xmark & \cmark & \xmark & \xmark \\
AIME~\citep{aime2025}                         & Math         & Rule   & \xmark & \xmark & \xmark & \xmark & \xmark & \cmark & \xmark \\
HMMT~\citep{hmmt2025}                         & Math         & Rule   & \xmark & \xmark & \xmark & \xmark & \xmark & \cmark & \xmark \\
USAMO~\citep{usamo2025}                       & Math         & Human  & \xmark & \xmark & \xmark & \xmark & \cmark & \cmark & \xmark \\
MMMLU~\citep{mmmlu2024}                       & Knowledge    & Rule   & \xmark & \xmark & \xmark & \xmark & \xmark & \xmark & \cmark \\
Video-MME~\citep{videomme2024}                 & Knowledge    & Rule   & \xmark & \cmark & \xmark & \xmark & \xmark & \xmark & \cmark \\
OpenAI-MRCR~\citep{mrcr2025}                  & Knowledge    & Rule   & \xmark & \xmark & \xmark & \xmark & \xmark & \xmark & \xmark \\
HLE~\citep{hle2025}                           & Knowledge    & Rule   & \xmark & \cmark & \xmark & \xmark & \cmark & \xmark & \cmark \\
ARC-AGI-2~\citep{arcagi2025}                  & Reasoning    & Rule   & \xmark & \xmark & \xmark & \xmark & \xmark & \xmark & \xmark \\
OODBench~\citep{oodbench2026}                  & Knowledge    & Rule   & \xmark & \cmark & \xmark & \xmark & \xmark & \xmark & \xmark \\
\midrule

\multicolumn{10}{l}{\cellcolor{coolblue3}\textbf{Agent Product Evaluation}} \\
\midrule
xbench~\citep{xbench2025}                        & Recruit+Mkt  & Model  & \cmark & \xmark & \xmark & \xmark & \cmark & \cmark & \xmark \\
AgentIF-OneDay~\citep{agentifoneday2026}          & Daily Tasks  & Model  & \xmark & \cmark & \xmark & \xmark & \cmark & \xmark & \cmark \\
\midrule

\multicolumn{10}{l}{\cellcolor{coolblue3}\textbf{Emerging Benchmarks (2026)}} \\
\midrule
Persona2Web~\citep{persona2web2026}            & Search       & Model  & \xmark & \xmark & \cmark & \xmark & \xmark & \xmark & \xmark \\
AmbiBench~\citep{ambibench2026}                & OS           & Model  & \xmark & \cmark & \cmark & \xmark & \xmark & \xmark & \xmark \\
PAHF~\citep{pahf2026}                         & Tool         & Rule+Model & \xmark & \xmark & \xmark & \xmark & \xmark & \xmark & \xmark \\
AgenticShop~\citep{agenticshop2026}            & Search       & Model  & \xmark & \xmark & \xmark & \xmark & \xmark & \xmark & \xmark \\
GAP Benchmark~\citep{gapbench2026}             & Tool         & Rule   & \xmark & \xmark & \xmark & \xmark & \xmark & \xmark & \cmark \\
AgentLAB~\citep{agentlab2026}                  & Safety         & Model  & \xmark & \xmark & \xmark & \xmark & \xmark & \xmark & \cmark \\
STING~\citep{sting2026}                       & Safety         & Model  & \xmark & \xmark & \xmark & \xmark & \xmark & \xmark & \xmark \\
GT-HarmBench~\citep{gtharmbench2026}           & Safety       & Rule   & \xmark & \xmark & \xmark & \xmark & \xmark & \xmark & \xmark \\
ForesightSafety~\citep{foresightsafety2026}    & Safety  & Model  & \xmark & \xmark & \xmark & \xmark & \xmark & \xmark & \cmark \\
APST~\citep{apst2026}                         & Safety     & Model  & \xmark & \xmark & \xmark & \xmark & \xmark & \xmark & \xmark \\
MemoryArena~\citep{memoryarena2026}            & Search+Research & Rule & \xmark & \xmark & \xmark & \xmark & \xmark & \xmark & \xmark \\
WebWorld-Bench~\citep{webworldbench2026}       & Search+Code  & Rule   & \xmark & \xmark & \xmark & \xmark & \xmark & \xmark & \cmark \\
Gaia2~\citep{gaia2_2026}                      & OS+Search    & Rule   & \xmark & \xmark & \xmark & \xmark & \xmark & \xmark & \xmark \\
SkillsBench~\citep{skillsbench2026}            & Code+Tool    & Rule   & \xmark & \xmark & \xmark & \xmark & \xmark & \xmark & \cmark \\
MATEO~\citep{mateo2026}                       & Reasoning    & Rule   & \xmark & \cmark & \xmark & \xmark & \xmark & \xmark & \xmark \\
SciAgentGym~\citep{sciagentbench2026}          & Research+Tool & Rule+Model & \xmark & \xmark & \xmark & \xmark & \xmark & \xmark & \cmark \\
Drug Scouting~\citep{drugscouting2026}         & Search+Research & Model & \xmark & \xmark & \xmark & \xmark & \xmark & \xmark & \xmark \\
AD-Bench~\citep{adbench2026}                  & Tool         & Rule   & \cmark & \xmark & \xmark & \xmark & \cmark & \xmark & \xmark \\
GUI-GENESIS~\citep{guigenesis2026}             & OS           & Rule   & \xmark & \cmark & \xmark & \xmark & \xmark & \xmark & \xmark \\
BookingArena~\citep{bookingarena2026}          & Search       & Rule+Model & \xmark & \xmark & \xmark & \xmark & \xmark & \xmark & \xmark \\
BrowseComp-V\textsuperscript{3}~\citep{browsecompv3_2026} & Search & Rule & \xmark & \cmark & \xmark & \xmark & \cmark & \xmark & \xmark \\
Collective Behavior~\citep{collectivebehavior2026} & Research & Rule  & \xmark & \xmark & \xmark & \xmark & \xmark & \xmark & \xmark \\
Unsafer~\citep{unsafer2026}                   & Tool         & Model  & \xmark & \xmark & \xmark & \xmark & \xmark & \xmark & \xmark \\
Proxy State Eval~\citep{proxystate2026}        & Tool         & Model  & \xmark & \xmark & \xmark & \xmark & \xmark & \xmark & \xmark \\

\midrule
\rowcolor{tablegreen}
\textbf{\modelname (Ours)} & \textbf{6 Domains} & \textbf{Rule+Model} & \cmark & \cmark & \cmark & \cmark & \cmark & \cmark & \cmark \\

\end{longtable}
\endgroup

\section{From Production Requirements to Executable Benchmarks}
\label{sec:pipeline}

\begin{figure}[H]
    \centering
    \includegraphics[width=\linewidth]{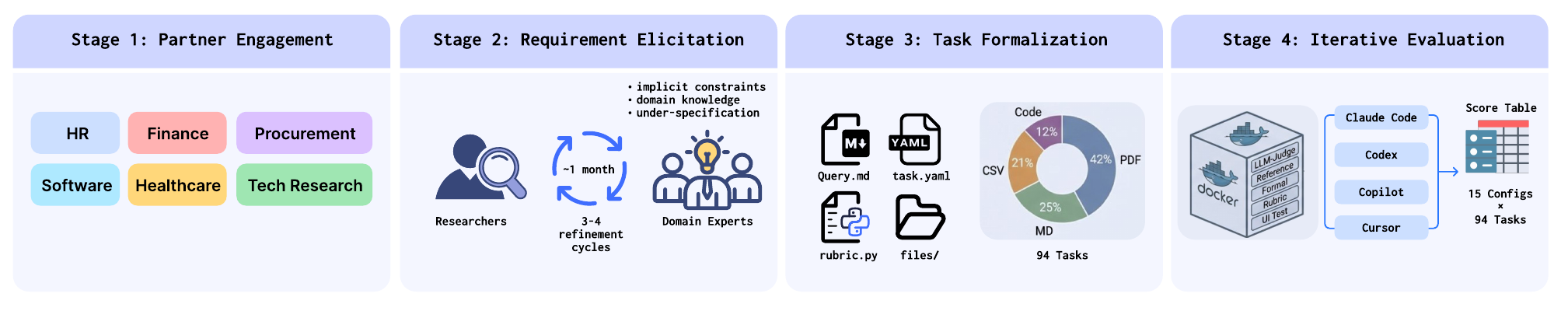}
    \caption{The requirement-to-benchmark construction framework: four stages from production requirements to automated evaluation.}
    \label{fig:pipeline}
\end{figure}

Existing benchmarks are constructed retrospectively: researchers select artifacts (e.g., resolved GitHub issues), then design evaluation criteria around them. This approach yields clean, reproducible tasks but fundamentally cannot capture the under-specification, implicit constraints, and domain expertise that characterize production work. \modelname inverts this direction: we start from \textit{authentic production requirements}---the actual tasks companies need AI agents to perform for paying customers---and systematically transform them into executable, automated evaluations.

This requirement-to-benchmark construction framework is itself a central contribution. The key challenge is not building a benchmark \textit{perse}, but providing a standardized, repeatable process from a real-world production need to a rigorous, reproducible evaluation. This enables evaluation of agents at a fundamentally deeper level---not just \textit{can the agent code?} but \textit{can the agent deliver value on the actual tasks businesses pay for?} Below we describe each stage of this framework (Figure~\ref{fig:pipeline}).

\paragraph{Partner Engagement.}
We partner with companies whose professional workflows intersect with AI agent capabilities, targeting diversity in O*NET occupational domains (Table~\ref{tab:companies}). Partners include both companies deploying AI agents as core products for paying customers and domain-expert organizations whose daily workflows generate authentic tasks suitable for agent evaluation (e.g., technology media producing in-depth industry analysis). We select partners that satisfy as many of the following criteria as possible: (1) access to authentic, professionally validated task requirements that yield long-horizon deliverables (complete reports, codebases, or analytical artifacts rather than short answers); (2) AI agents integral to revenue-generating workflows; (3) diverse input modalities; (4) domain expertise to co-design stakeholder-aligned evaluation criteria and iteratively refine them as business needs evolve; (5) willingness to share anonymized data.

\begin{table}[t]
\centering
\resizebox{\linewidth}{!}{
\begin{tabular}{llccl}
\toprule
\textbf{Domain (O*NET)} & \textbf{Representative Task} & \textbf{Tasks} & \textbf{Input} & \textbf{Eval Paradigm} \\
\midrule
Human Resources (13-1071) & Resume screening vs.\ JD & 11 & PDF, JPEG & F1 score \\
Finance \& Investment (13-2051) & Segment research \& pitch critique & 22 & PDF, MD, TXT & LLM-as-a-Judge \\
Procurement \& Operations (13-1020) & BOM cost optimization & 23 & Excel, CSV, MD & Constraint verif. \\
Software Engineering (15-1252) & Full-stack app generation & 11 & YAML, Code & UI testing \\
Healthcare \& Life Sci. (29-9099) & eCRF \& insurance policy analysis & 16 & PDF, MD & LLM + Numerical \\
Technology Research (15-1221) & AI industry deep analysis & 11 & MD & LLM-as-a-Judge \\
\midrule
\textbf{Total} & & \textbf{94} & & \\
\bottomrule
\end{tabular}
}
\caption{Overview of \modelname task domains classified by O*NET occupational taxonomy~\citep{onetonline}, where Eval Paradigm indicates the primary evaluation method.}
\label{tab:companies}
\end{table}

\paragraph{Requirement Elicitation.}
For each company, we conduct structured engagement spanning approximately one month, combining both online meetings and on-site visits. The critical insight is that production requirements rarely arrive as well-specified task descriptions---they emerge through iterative dialogue. A typical engagement proceeds through three phases: (1)~\textit{Workflow discovery}---companies demonstrate their end-to-end workflows, often revealing that the actual task complexity far exceeds initial descriptions (e.g., a clinical data partner initially described their need as ``converting Word documents to JSON,'' but meetings revealed a four-layer reasoning chain: temporal phase identification, trigger rule extraction, form field mapping, and constraint validation); (2)~\textit{Scope negotiation}---we jointly determine which segments of long production pipelines can be isolated into self-contained evaluation tasks while remaining professionally meaningful (e.g., negotiating whether to evaluate the full end-to-end CRF construction or focus on visit window computation, which captures the core reasoning challenge); and (3)~\textit{Ground truth co-construction}---domain experts provide or validate reference outputs, as standard answers often do not exist in pre-defined form and must be extracted from actual business decisions (e.g., using real interview shortlists rather than AI-generated candidate rankings). Through at least weekly meetings, we jointly develop task specifications that preserve the original level of under-specification, implicit constraints, and domain knowledge dependency that make production tasks fundamentally harder than research tasks.

\paragraph{Task Formalization.}
We provide partner companies with our standardized task package format upfront, and they deliver their tasks according to this structure---ensuring consistency across domains while allowing domain-specific flexibility in evaluation design. Each task is formalized into a self-contained package: (1) \textbf{Task Specification} (\texttt{query.md})---natural language description preserving the original level of specification; (2) \textbf{Task Configuration} (\texttt{task.yaml})---structured metadata including task name, domain category, difficulty level, evaluation type, and agent timeout; (3) \textbf{Input Files} (\texttt{files/})---raw documents required by the task (PDFs, Excel, images, etc.); (4) \textbf{Evaluation Specification} (\texttt{.eval/rubric.py})---composing one or more evaluation paradigms, with optional \texttt{ground\_truth.json} as reference answers. This format serves as a shared contract between our team and partners: we define the structure, partners populate it with authentic production content. Approximately 42\% of tasks involve PDFs, 21\% structured data files, 25\% markdown/text, and 12\% code/YAML.

\paragraph{Iterative Validation.}
We validate evaluation tasks using frontier agents internally and through collaborative verification with partner companies, averaging three to four refinement cycles per company. This iterative process ensures stakeholder-aligned evaluation: rubrics faithfully capture the quality dimensions that matter to paying customers---not just technical correctness, but the holistic deliverable quality that determines whether a business would accept the agent's output. Crucially, these criteria are not static: as agent capabilities improved during our evaluation period, several partners raised their quality bars, reflecting the dynamic nature of production evaluation standards.

\section{The \modelname Benchmark}
\label{sec:benchmark}

\modelname comprises \textbf{94} tasks sourced from companies with paying customers, classified into six O*NET occupational domains (summary statistics in Table~\ref{tab:benchmark_stats}). Several characteristics distinguish it from existing benchmarks: \textit{production provenance} (every task from an active commercial deployment), \textit{implicit constraints} (requirements contain hidden rules and undeclared priorities invisible to outsiders), \textit{information fragmentation} (key information is scattered across different locations in multiple documents, requiring cross-document reasoning), \textit{domain knowledge dependency} (tasks demand expertise beyond the task specification, such as medical insurance policies, investment analysis frameworks, and GCP standards), \textit{multi-modal heterogeneity} (agents process PDFs, spreadsheets, scanned images, and code within single tasks), \textit{long-horizon deliverables} (outputs are complete professional artifacts---10-page investment reports, full-stack codebases, multi-visit clinical calculations---rather than short answers or code snippets), \textit{stakeholder-aligned and evolving evaluation} (evaluation criteria are co-designed with domain practitioners who represent actual customers, and may shift as business requirements and agent capabilities evolve), and \textit{evaluation pluralism} (covers multiple evaluation paradigms).

\paragraph{Task Categories.}
We classify tasks following the O*NET occupational taxonomy~\citep{onetonline}, ensuring each domain maps to a standardized work activity category. This classification enables systematic coverage analysis and makes the benchmark extensible---any new production task can be mapped to an existing O*NET domain.

\textit{Human Resources} (11 tasks, O*NET 13-1071): Agents screen candidate resumes against job descriptions. A representative task provides 24 resumes (PDF and JPEG) for an AI Scientist internship at a startup accelerator, mixing objective criteria (``at least one internship longer than 3 months'') with subjective ones (``passion for cutting-edge technology''). The agent must select exactly 6 candidates; evaluation computes F1 against actual interview decisions.

\textit{Finance \& Investment} (22 tasks, O*NET 13-2051): Agents perform investment research and financial data extraction. Tasks include: (a)~generating professional segment research reports from startup business plans, following prescribed templates covering market sizing (TAM-SOM-SAM), competitive landscape, and technology deep-dive (14 tasks); (b)~synthesizing unstructured meeting transcripts into actionable investor critiques---e.g., designing a 2-minute pitch storyline and assessing whether a healthcare platform's 20,000-user reach translates to genuine product-market fit given only 70 core users (5 tasks); and (c)~extracting structured financial data from multi-year corporate annual reports (3 tasks). Evaluation combines LLM-as-a-Judge with structural validation.

\textit{Procurement \& Operations} (23 tasks, O*NET 13-1020): Agents solve constrained optimization and data processing problems grounded in real procurement workflows. A representative task presents 2,000 board cards with specifications and pricing in Excel, plus a natural language requirements document with implicit constraints (20 tasks). Additional tasks involve procurement bidding data analysis and spreadsheet operations (3 tasks). Evaluation programmatically verifies constraint satisfaction and cost optimality.

\textit{Software Engineering} (11 tasks, O*NET 15-1252): Agents build or modify full-stack mobile applications from detailed product requirements. A representative task provides database schemas, seed data, and a 200-line requirements document for a poetry appreciation app (built with the UniApp framework) featuring AI-powered playback, community posting, and collection management across four navigation pages. Evaluation employs automated end-to-end UI testing: a headless browser executes user flows and scores 10+ functional requirements.

\textit{Healthcare \& Life Sciences} (16 tasks, O*NET 29-9099): Agents handle clinical trial management and healthcare policy analysis. Tasks include: (a)~computing visit windows for electronic Case Report Form (eCRF) systems, where shifting one visit propagates through subsequent windows via cascade dependencies (10 tasks); and (b)~analyzing pharmaceutical reimbursement policies, medical insurance calculations, and drug coverage regulations (6 tasks). Evaluation combines LLM-as-a-Judge with numerical verification.

\textit{Technology Research} (11 tasks, O*NET 15-1221): Agents conduct comprehensive investigations on current technology topics, requiring web search, multi-source synthesis, and structured report generation. A representative task asks the agent to research the survival status of AI agent startups that raised over \$100M, including failures, acquisitions, and emerging trends. Evaluation uses LLM-as-a-Judge with weighted rubric points requiring source-backed evidence.


\paragraph{Evaluation Methodologies.}
\modelname covers four of the five major evaluation paradigms identified in our taxonomy (Section~\ref{sec:background}): \textbf{Reference Answer Verification}, \textbf{Formal Logic Verification}, \textbf{Rubric-based Evaluation}, and \textbf{Execution-based Verification} (detailed per-paradigm coverage in Table~\ref{tab:eval_coverage}). We additionally employ LLM-as-a-Judge (Claude Opus 4.6) as a cross-cutting semantic evaluation method within multiple paradigms. Critically, individual domains \textit{compose} multiple paradigms---e.g., clinical research combines LLM-as-a-Judge with numerical verification. Every task employs $\geq$2 leaf-node evaluation types (avg.\ 2.8; Table~\ref{tab:eval_task_composition}), reflecting the multi-dimensional nature of production quality. All rubric scripts output standardized scores (0.0--1.0) enabling cross-domain comparison.

\paragraph{Economic Value Annotation.}
To ground task difficulty in economic terms, we annotate each task with its \textit{human replacement cost} through a two-stage pipeline: automated AI estimation followed by domain expert calibration (full methodology in Appendix~\ref{sec:appendix_value}). After calibration, the 94 tasks represent \textbf{2,420 professional hours} ($\sim$60 person-weeks) of labor, valued at \textbf{\$154K--\$231K} (USD) or \textbf{\textyen391K--\textyen570K} (CNY)---validating that \modelname captures economically meaningful work.

\paragraph{Evaluation Infrastructure.}
\label{sec:infrastructure}
The \modelname framework is organized around three abstractions: a \textbf{Task Runner} managing evaluation lifecycle, an \textbf{Evaluator Registry} routing tasks to paradigm-specific pipelines, and an \textbf{Execution Sandbox} using Docker containers for isolation. Each rubric script outputs a standardized score $s_{\text{task}} = \sum_{k} w_k \cdot e_k \in [0, 1]$, where $e_k$ is the evaluation result from paradigm $k$ with expert-assigned weight $w_k$. Domain scores are unweighted task means; the overall score is the unweighted mean across six domains, giving each domain equal influence regardless of task count (see Appendix~\ref{sec:appendix_scoring} for details).

\paragraph{Challenges in Production Evaluation.}
\label{sec:challenges}
Constructing \modelname surfaced challenges largely absent from research benchmark development: (1)~\textit{Preserving productive ambiguity}---fully disambiguating tasks destroys what makes production tasks difficult; (2)~\textit{Implicit constraints}---practitioners consider requirements ``obvious'' that are absent from written specifications; (3)~\textit{Quantifying subjective judgment}---experts assess quality holistically, and decomposing into verifiable dimensions inevitably loses information; (4)~\textit{Evaluation criteria drift}---because evaluation is stakeholder-aligned, partner companies revised quality standards as agent capabilities improved and business priorities shifted, requiring ongoing rubric maintenance; (5)~\textit{Environment fidelity}---production agents operate within rich software environments difficult to reproduce in sandboxed evaluation; and (6)~\textit{Balancing openness and confidentiality}---partner companies are cautious about sharing evaluation criteria encoding competitive knowledge.

\section{Experiments}
\label{sec:experiments}

\subsection{Setup}

\paragraph{Models.}
We evaluate six frontier models spanning both closed-source and open-source ecosystems. \textit{Closed-source models}: Claude Opus 4.6 (Anthropic), GPT-5.2 (OpenAI), and Gemini 3 Pro Preview (Google)---representing the strongest proprietary offerings from three major AI labs. \textit{Open-source models}: Kimi K2.5 (Moonshot), GLM-5 (Zhipu AI), and MiniMax M2.5 (MiniMax)---representing competitive open-weight alternatives. This selection enables direct comparison of the open-source vs.\ closed-source performance gap in production settings.

\paragraph{Agent Systems.}
Models are deployed through four commercial agent products: \textbf{Claude Code} (Anthropic), \textbf{Codex} (OpenAI), \textbf{GitHub Copilot} (GitHub), and \textbf{Cursor} (Cursor). All agents are invoked via their respective CLI interfaces within Docker-sandboxed environments with pinned versions for reproducibility (Table~\ref{tab:agent_config} in Appendix). We record full output trajectories (tool calls, intermediate reasoning, and final artifacts) for each run, enabling post-hoc failure analysis.

\paragraph{Configurations.}
While $6 \times 4 = 24$ combinations are theoretically possible, we evaluate 14 configurations (Table~\ref{tab:overall_results}), selecting combinations based on two criteria: (1)~\textit{real-world adoption}---we prioritize model--scaffold pairings that are widely used in production settings by our partner companies and research teams (e.g., Claude Code + Opus, Codex + GPT-5.2, Cursor + Opus); and (2)~\textit{evaluation cost}---each full benchmark run involves 94 tasks with substantial agent execution (e.g., Claude Code + Opus averages 46 turns and 14 minutes per task), making exhaustive enumeration prohibitively expensive. The selected 14 configurations cover all four scaffolds and all six models, ensuring comprehensive coverage while focusing on practically relevant pairings.

\subsection{Results and Domain-Level Analysis}

\begin{table*}[t]
\centering
\tiny
\resizebox{\textwidth}{!}{
\begin{tabular}{ll|cccccc|c|c}
\toprule
\textbf{Agent Product} & \textbf{Model} & \textbf{HR} & \textbf{F\&I} & \textbf{P\&O} & \textbf{SE} & \textbf{H\&LS} & \textbf{TR} & \textbf{Avg.} & \textbf{Value (USD K)} \\
 & & (11) & (22) & (23) & (11) & (16) & (11) & & \\
\midrule
\multirow{6}{*}{Claude Code}
& Claude Opus 4.6 & \underline{35.91} & \textbf{70.35} & 83.35 & \textbf{70.95} & \textbf{50.06} & \underline{75.82} & \textbf{64.41} & \textbf{110--165} \\
& Gemini 3 Pro & 26.73 & 54.68 & 65.39 & 59.47 & 39.88 & 58.55 & 50.78 & 89--133 \\
& GPT-5.2 & 26.55 & 38.64 & 40.22 & 55.08 & 27.94 & 48.36 & 39.47 & 70--106 \\
& Kimi K2.5 & 20.18 & 47.80 & 35.04 & 56.77 & 44.44 & 59.18 & 43.90 & 78--116 \\
& GLM-5 & 31.91 & 51.18 & 66.43 & 52.46 & 38.88 & 51.36 & 48.70 & 82--122 \\
& MiniMax M2.5 & 26.00 & 52.55 & 39.39 & 42.03 & 34.62 & 50.73 & 40.89 & 70--105 \\
\midrule
\multirow{4}{*}{Codex}
& GPT-5.2 & 27.36 & 62.80 & 52.35 & 50.56 & 32.19 & 60.27 & 47.59 & 84--126 \\
& GLM-5 & 24.36 & 57.85 & 56.74 & 53.58 & \underline{46.69} & 59.91 & 49.85 & 85--128 \\
& Claude Opus 4.6 & 32.18 & 45.99 & 79.00 & 54.74 & 36.81 & 72.00 & 53.45 & 86--129 \\
& Kimi K2.5 & 35.36 & 47.27 & 30.91 & 52.52 & 38.75 & 53.73 & 43.09 & 73--109 \\
\midrule
\multirow{3}{*}{GitHub Copilot}
& GPT-5.2 & 32.00 & 63.39 & 74.04 & 61.07 & 30.69 & 68.27 & 54.91 & 97--145 \\
& Gemini 3 Pro & 28.27 & 59.31 & 65.35 & 50.46 & 30.69 & 65.45 & 49.92 & 86--128 \\
& Claude Opus 4.6 & 34.18 & 62.12 & \textbf{88.09} & 63.07 & 44.06 & \textbf{76.36} & 61.31 & 102--153 \\
\midrule
Cursor
& Claude Opus 4.6 & \textbf{38.91} & \underline{67.10} & \textbf{88.09} & \underline{64.77} & 44.06 & 68.18 & \underline{61.85} & \underline{104--156} \\
\bottomrule
\end{tabular}
}
\caption{Average scores (0--100) on \modelname across six O*NET domains. Value (USD K) estimates the human replacement cost each configuration delivers, computed as $\sum_d \text{score}_d \times \text{domain\_value}_d$ using per-domain expert-calibrated costs (Table~\ref{tab:value_summary}). \textbf{Bold}: best in column. \underline{Underline}: second best.}
\label{tab:overall_results}
\end{table*}

Several key findings emerge from Table~\ref{tab:overall_results}:

\begin{itemize}[leftmargin=*,itemsep=2pt,topsep=2pt]
    \item \textbf{Low absolute scores.} The best configuration (Claude Code + Opus 4.6) achieves only 64.41 on average, revealing a substantial research-production gap.
    \item \textbf{Scaffold matters as much as model.} The same Claude Opus 4.6 scores 64.41 via Claude Code, 61.85 via Cursor, 61.31 via GitHub Copilot, but only 53.45 via Codex. GPT-5.2 scores 39.47 via Claude Code but 54.91 via GitHub Copilot---a 15-point spread. This confirms that evaluating \textit{agent systems}, not just models, is essential.
    \item \textbf{Extreme domain variance.} Procurement \& Operations scores range from 30.91 to 88.09, while Human Resources scores never exceed 38.91---no single aggregate score captures production readiness. Model rankings are also domain-dependent: GLM-5 scores 66.43 on Procurement \& Operations but only 52.46 on Software Engineering, meaning aggregate rankings would be misleading.
    \item \textbf{Score ranking $\neq$ value ranking: implications for agent selection.} Domain-weighted economic value reveals a different picture from average scores. Codex + Opus 4.6 (avg.\ 53.45, \$86K--\$129K) outscores Claude Code + Gemini 3 Pro (avg.\ 50.78) but delivers \textit{less} economic value (\$89K--\$133K), because Gemini 3 Pro performs better on high-value domains (Software Engineering, Finance \& Investment). This has direct implications for agent selection in practice: (1)~organizations should select configurations based on their \textit{domain portfolio}---a company primarily doing financial analysis should weight F\&I performance heavily, not rely on aggregate scores; (2)~a \textit{multi-agent strategy} may be optimal, routing different task types to different configurations (e.g., Claude Code + Opus for finance and research, Copilot + Opus for procurement); and (3)~the \$40K--\$60K value gap between configurations provides organizations with a quantitative basis for agent selection decisions, grounding what is typically an intuition-driven choice in concrete economic terms.
\end{itemize}

We order domains from easiest to hardest (highest to lowest average score across configurations): \textbf{Technology Research} (avg.~62.0): scores range from 48.36 to 76.36, with strong performance from configurations using Opus 4.6. Tasks requiring up-to-date information retrieval, multi-source synthesis, and technical depth remain challenging, but agents with persistent search strategies achieve meaningful scores.
\textbf{Procurement \& Operations} (avg.~61.7): Cursor + Opus 4.6 and Copilot + Opus 4.6 reach 88.09; binary pass/fail scoring on the core optimization tasks reflects the zero-tolerance nature of procurement decisions.
\textbf{Software Engineering} (avg.~56.3): even weaker models exceed 42 points, but the gap between top (70.95) and bottom (42.03) remains substantial.
\textbf{Finance \& Investment} (avg.~55.8): Claude Code + Opus 4.6 leads at 70.35, with strong performance from Codex configurations that achieve near-perfect scores on financial data extraction subtasks.
\textbf{Healthcare \& Life Sciences} (avg.~38.6): Claude Code + Opus 4.6 leads at 50.06; clinical trial eCRF tasks exhibit zero tolerance for numerical errors, while healthcare policy tasks require domain-specific regulatory knowledge.
\textbf{Human Resources} (avg.~30.0): best score only 38.91---aligning agent judgments with human hiring decisions remains difficult.

\paragraph{Economic Value Delivered.}
Translating scores into economic terms using per-domain expert-calibrated labor costs (Appendix~\ref{sec:appendix_value}, Table~\ref{tab:value_summary}), the best configuration delivers an estimated \textbf{\$110K--\$165K} in professional labor value while the worst delivers \$70K--\$105K---a \textbf{\$40K--\$60K gap} from the same 94 tasks. The Value column in Table~\ref{tab:overall_results} reports these domain-weighted values for all 14 configurations, providing organizations with a quantitative basis for agent selection beyond aggregate scores alone.

\subsection{Evaluation Reliability}

\paragraph{Statistical Reliability.}
To address single-run stochasticity, we conduct repeated evaluations for our best-performing configuration (Claude Code + Opus 4.6) across three independent runs (Table~\ref{tab:ci}). The narrow confidence intervals (overall $\pm$1.83) confirm that reported scores are stable and that configuration rankings are reproducible across runs. Variance naturally differs across evaluation paradigms---constraint-verification domains show higher variance (P\&O std=4.72) than LLM-as-a-Judge domains (F\&I std=1.87, TR std=3.56)---reflecting the inherent characteristics of each paradigm rather than evaluation instability.

\begin{table}[h]
\centering
\small
\begin{tabular}{@{}lccc@{}}
\toprule
\textbf{Domain} & \textbf{Mean} & \textbf{Std} & \textbf{95\% CI} \\
\midrule
Human Resources & 35.91 & 2.14 & [33.77, 38.05] \\
Finance \& Investment & 70.35 & 1.87 & [68.48, 72.22] \\
Procurement \& Ops. & 83.35 & 4.72 & [78.63, 88.07] \\
Software Engineering & 70.95 & 3.21 & [67.74, 74.16] \\
Healthcare \& Life Sci. & 50.06 & 2.93 & [47.13, 52.99] \\
Technology Research & 75.82 & 3.56 & [72.26, 79.38] \\
\midrule
\textbf{Overall} & \textbf{64.41} & \textbf{1.83} & \textbf{[62.58, 66.24]} \\
\bottomrule
\end{tabular}
\caption{Repeated evaluation results for Claude Code + Opus 4.6 (3 runs). CI = 95\% confidence interval.}
\label{tab:ci}
\end{table}

\paragraph{Meta-Evaluation.}
We validate evaluation reliability on 20 randomly sampled LLM-as-a-Judge tasks across all six domains (5 configurations, 1,000 rubric point judgments). Two independent expert annotators (strict A, lenient B) assess each rubric point alongside the automated judge (Table~\ref{tab:meta_eval}).

\begin{table}[h]
\centering
\small
\begin{tabular}{@{}lcccc@{}}
\toprule
\textbf{Pair} & \textbf{Agr.} & \textbf{$\kappa$} & \textbf{$\rho$} & \textbf{$r$} \\
\midrule
A vs.\ B & 84.7\% & 0.691 & 0.818 & 0.870 \\
A vs.\ LLM-as-a-Judge & 85.0\% & 0.697 & 0.820 & 0.861 \\
B vs.\ LLM-as-a-Judge & 89.7\% & 0.780 & 0.845 & 0.885 \\
\midrule
Three-way (Fleiss) & 79.7\% & 0.720 & --- & --- \\
\bottomrule
\end{tabular}
\caption{Meta-evaluation results: pairwise agreement between two expert annotators (A, B) and the automated LLM-as-a-Judge evaluation across 1,000 rubric point judgments.}
\label{tab:meta_eval}
\end{table}

All pairwise Cohen's $\kappa$ values fall within the \textit{substantial agreement} range (0.69--0.78), with Fleiss' $\kappa = 0.720$ confirming three-way reliability. The automated judge shows higher agreement with the lenient annotator ($\kappa = 0.780$) than the strict one ($\kappa = 0.697$), consistent with known LLM-as-a-Judge evaluation biases such as self-preference~\citep{panickssery2024} and self-enhancement~\citep{zheng2023judging}.

\subsection{Failure Mode Analysis}

Beyond aggregate performance gaps, we conduct qualitative analysis to understand \textit{why} agents fail on production tasks. We identify six production-specific failure modes invisible to coding benchmarks (detailed case studies in Appendix~\ref{sec:appendix_error} and~\ref{sec:appendix_crossdomain}):

\textbf{(1) Cascade dependency failure.} In Healthcare \& Life Sciences, misidentifying a Day~1 anchor produces systematically incorrect window calculations for all subsequent visits. In Finance \& Investment, misidentifying a company's industry cascades through market sizing, comparable company selection, and competitive analysis.

\textbf{(2) Subjective judgment collapse.} In Human Resources, agents extract factual qualifications but fail on soft-skill inference. Tasks with quantifiable criteria score 2--3$\times$ higher than those requiring holistic judgment.

\textbf{(3) Information retrieval failures.} Technology Research tasks expose five cognitive failure modes: factual hallucination ($\sim$30\%), imprecise retrieval ($\sim$35\%), rigid search strategies ($\sim$15\%), attribution confusion ($\sim$10\%), and positive-information bias ($\sim$10\%). For example, Claude Opus 4.6 substitutes outdated Series~D data for Harvey AI's F-round with every data point incorrect by a large margin and no hedging language (Table~\ref{tab:hallucination} in Appendix). Models also systematically miss negative events (startup failures, funding collapses) because default searches surface success stories---\textit{no model} reported Robin AI's distressed sale despite it being a prominent Series~C failure.

\textbf{(4) Cross-section logical inconsistency.} In Finance \& Investment, agents produce individually plausible paragraphs that contradict each other---a TAM of \$50B in one section but \$80B two pages later. Models lack a global coherence mechanism across long-form outputs.

\textbf{(5) Constraint misinterpretation.} In Procurement \& Operations, agents optimize explicitly stated objectives while violating implicit constraints, and exhibit ``synergy blindness''---optimizing components independently rather than jointly. When a procurement problem has no feasible solution (conflicting constraints), the majority of agent responses fabricate a ``best effort'' solution rather than declaring infeasibility---a particularly dangerous behavior in production.

\textbf{(6) Format compliance failures.} The most production-specific failure: agents produce substantively reasonable analyses that score poorly because the output format is incompatible with downstream consumption.

\section{Discussion}
\label{sec:discussion}

\modelname demonstrates that production-grounded evaluation reveals capability gaps invisible to research benchmarks---not merely harder tasks, but a qualitative mismatch between the skills research benchmarks select for (precise instruction following, deterministic reasoning, short-horizon outputs) and what production demands (tolerance for ambiguity, domain-appropriate judgment, long-horizon deliverables, and format compliance under stakeholder-defined quality standards). For model developers, \modelname identifies failure modes invisible until agents encounter real business requirements; for deploying organizations, it offers ready-made evaluation infrastructure. We open-source the evaluation framework and construction methodology, enabling the community to build production-grounded benchmarks for their own domains---addressing the continuous evolution challenge through collaborative development.

\paragraph{From Capability Measurement to Value Measurement.}
A distinctive feature of \modelname is its economic value grounding, which enables a shift from asking ``how well does the agent perform?'' to ``how much value does the agent deliver?'' Our domain-weighted analysis reveals that the best agent configuration delivers an estimated \$110K--\$165K in equivalent professional labor value across the benchmark, while the worst delivers only \$70K--\$105K. This framing has practical implications: organizations can directly compare agent licensing costs against the value differential between configurations, making agent selection an economic optimization problem rather than a purely technical one.

\paragraph{Limitations.}
Several limitations should be acknowledged. First, the current benchmark covers six O*NET domains from seven companies; while diverse, it does not yet span all occupational categories where agents are deployed (e.g., legal, education, creative industries). Second, our evaluation relies on a single snapshot in time---both agent capabilities and partner quality standards evolve, and longitudinal tracking remains future work. Third, the economic value estimates, while expert-calibrated, involve assumptions about benefit multipliers and wage distributions that may not generalize across all markets. Fourth, we evaluate only four commercial agent scaffolds; emerging open-source frameworks and custom enterprise pipelines are not yet represented. Finally, the 14 configurations, while covering all models and scaffolds, do not exhaust all possible pairings, and performance on untested combinations may differ.

\section{Conclusion}
We present \modelname, a production-grounded benchmark of 94 tasks sourced from seven companies across six O*NET occupational domains, together with a standardized requirement-to-benchmark construction framework. Our evaluation of 14 model--scaffold configurations across six frontier models and four commercial agent products reveals three key insights: (1)~the best configuration (Claude Code + Opus 4.6) achieves only 64.41/100, exposing a substantial gap between research benchmark performance and production readiness; (2)~scaffold choice matters as much as model choice---the same model can vary by 11--15 points depending on the agent product, confirming that evaluating complete \textit{agent systems} rather than models alone is essential; and (3)~domain-weighted economic value reveals that score rankings do not always align with value rankings, providing organizations with a more nuanced basis for agent selection. We identify six production-specific failure modes---cascade dependencies, subjective judgment collapse, information retrieval failures, cross-section inconsistency, constraint misinterpretation, and format compliance---that are invisible to coding-centric benchmarks. By open-sourcing the evaluation framework and construction methodology, we aim to enable community-driven evolution: organizations can adopt our standardized pipeline to build production-grounded benchmarks for their own domains.

\section*{Acknowledgments}
We thank Keyu Li for his valuable comments, and Tianze Xu and Zhen Huang for their valuable contributions to the early stages of this project.

\bibliography{bib}
\bibliographystyle{acl_natbib}

\appendix


\section{Evaluation Infrastructure Details}
\label{sec:appendix_infra}

The evaluation pipeline proceeds in six stages: (1) \textbf{Task Loading}: the framework scans the benchmark directory and constructs a task queue filterable by domain, difficulty, or evaluation type; (2) \textbf{Environment Provisioning}: Docker containers are instantiated with task input files; (3) \textbf{Agent Invocation}: the agent receives \texttt{query.md} and interacts through a standardized interface; (4) \textbf{Output Collection}: artifacts are collected from the results directory; (5) \textbf{Evaluation Dispatch}: outputs are routed to the appropriate evaluator; (6) \textbf{Result Aggregation}: results are aggregated into domain-level and benchmark-level summaries.

All task inputs, evaluation scripts, and reference answers are version-controlled. Docker images and agent scaffold versions are pinned for reproducibility (Claude Code 2.1.70, Codex 0.80.0/0.111.0, GitHub Copilot 1.0.10, Cursor 2026.03.11; see Table~\ref{tab:agent_config} for details). Each run generates structured logs with full interaction traces, raw outputs, and scores in JSON format.

\section{Scoring and Aggregation Methodology}
\label{sec:appendix_scoring}

\paragraph{Per-Task Scoring.}
Each task's evaluation rubric script (\texttt{rubric.py}) outputs a standardized score $s \in [0, 1]$, which is scaled to $[0, 100]$ for reporting. For tasks composing multiple evaluation paradigms, the rubric script internally handles the composition---typically as a weighted average of sub-evaluations, where weights are co-designed with domain experts during the task formalization stage. The final per-task score thus captures multi-dimensional quality in a single number:
\begin{equation}
s_{\text{task}} = \sum_{k=1}^{K} w_k \cdot e_k(a, t), \quad \text{where } \sum_{k=1}^{K} w_k = 1
\end{equation}
where $e_k(a, t) \in [0, 1]$ is the evaluation result from paradigm $k$ for agent $a$ on task $t$, and $w_k$ is the expert-assigned weight for paradigm $k$.

\paragraph{Domain-Level Aggregation.}
Domain scores are computed as unweighted arithmetic means across all tasks within a domain:
\begin{equation}
S_{\text{domain}}(a) = \frac{1}{|T_d|} \sum_{t \in T_d} s_{\text{task}}(a, t) \times 100
\end{equation}
where $T_d$ is the set of tasks in domain $d$.

\paragraph{Overall Aggregation.}
The overall benchmark score is the unweighted arithmetic mean across the six domain scores, giving each domain equal weight regardless of task count:
\begin{equation}
S_{\text{overall}}(a) = \frac{1}{6} \sum_{d=1}^{6} S_{\text{domain}}(a)
\end{equation}
This equal-weighting ensures that domains with fewer tasks (e.g., Human Resources with 11 tasks) receive the same influence as domains with more tasks (e.g., Procurement \& Operations with 23 tasks), reflecting the principle that production readiness requires competence across all occupational domains.

\section{Agent System Configuration}
\label{sec:appendix_agent_config}

All agent systems are invoked via their respective command-line interfaces (CLIs) within Docker-sandboxed environments. Table~\ref{tab:agent_config} summarizes the configuration details.

\begin{table}[h]
\centering
\small
\begin{tabular}{@{}llll@{}}
\toprule
\textbf{Agent Product} & \textbf{Version} & \textbf{Interface} & \textbf{Models} \\
\midrule
Claude Code & 2.1.70 & \texttt{claude} CLI & All 6 models \\
Codex & 0.111.0 & \texttt{codex} CLI & GPT-5.2, Opus 4.6 \\
Codex & 0.80.0 & \texttt{codex} CLI & GLM-5, Kimi K2.5 \\
GitHub Copilot & 1.0.10 & \texttt{copilot-cli} & GPT-5.2, Gemini 3 Pro, Opus 4.6 \\
Cursor & 2026.03.11 & \texttt{cursor} CLI & Opus 4.6 \\
\bottomrule
\end{tabular}
\caption{Agent system configurations with pinned versions. Codex version varies by model due to API compatibility requirements.}
\label{tab:agent_config}
\end{table}
\paragraph{Trajectory Recording.}
Each evaluation run produces a complete execution trajectory capturing: (1) all tool calls and their arguments, (2) intermediate reasoning steps (where exposed by the scaffold), (3) file read/write operations, and (4) the final output artifacts. These trajectories enable post-hoc failure analysis---for instance, identifying whether a low score resulted from the model failing to find relevant information (search failure) or finding it but incorporating it incorrectly (reasoning failure). Although some agent products are closed-source, the CLI interface ensures that the agent's observable behavior---tool calls, outputs, and timing---is fully recorded and reproducible.

\paragraph{Model Configuration.}
All models are evaluated using their default API configurations (temperature, max tokens, etc.) as provided by each scaffold. We do not modify model parameters, as the goal is to evaluate agent \textit{products} as deployed, not to optimize individual model settings. This mirrors how production users interact with these systems.

\section{Evaluation Taxonomy Coverage}
\label{sec:appendix_taxonomy}

Table~\ref{tab:eval_coverage} maps the leaf-node evaluation types from our taxonomy (Figure~\ref{fig:eval_taxonomy}) to the domains that employ them. \modelname covers 8 of the 14 leaf-node types across four paradigms; the uncovered types (word-embedding similarity, repo integration testing, auto-generated rubrics, human-AI collaborative rubrics, and pairwise ranking) are either inapplicable to production tasks or reserved for future expansion.

\begin{table}[h]
\centering
\small
\setlength{\tabcolsep}{4pt}
\begin{tabular}{@{}llccl@{}}
\toprule
\textbf{Paradigm} & \textbf{Leaf-node Type} & \textbf{Covered} & \textbf{Tasks} & \textbf{Domains} \\
\midrule
\multirow{4}{*}{\shortstack[l]{Reference\\Verif.}}
  & Fuzzy matching & \cmark & 42 & HR, F\&I, P\&O \\
  & Exact matching & \cmark & 33 & HR, F\&I, P\&O \\
  & Word-embed.\ sim. & --- & 0 & --- \\
  & LLM semantic & \cmark & 49 & F\&I, H\&LS, TR \\
\midrule
\multirow{3}{*}{\shortstack[l]{Test Case\\Verif.}}
  & Code unit testing & \cmark & 20 & P\&O \\
  & Repo integ.\ test & --- & 0 & --- \\
  & Env.\ state verif. & \cmark & 13 & F\&I, SE \\
\midrule
\multirow{2}{*}{\shortstack[l]{Formal\\Verif.}}
  & Math proof & \cmark & 25 & H\&LS, F\&I, P\&O \\
  & Code/logic & \cmark & 2 & H\&LS \\
\midrule
\multirow{3}{*}{\shortstack[l]{Rubric\\Assess.}}
  & Human-authored & \cmark & 49 & F\&I, H\&LS, TR \\
  & Auto-generated & --- & 0 & --- \\
  & Human-AI collab. & --- & 0 & --- \\
\midrule
Competition & Ranking & --- & 0 & --- \\
\bottomrule
\end{tabular}
\caption{Fine-grained evaluation type coverage across the paradigms implemented by \modelname.}
\label{tab:eval_coverage}
\end{table}

\section{Benchmark Statistics}
\label{sec:appendix_stats}

Table~\ref{tab:benchmark_stats} summarizes the key statistics of \modelname.

\begin{table}[h]
\centering
\begin{tabular}{lr}
\toprule
\textbf{Statistic} & \textbf{Value} \\
\midrule
Total tasks & 94 \\
Domains covered & 6 \\
Partner companies & 7 \\
\midrule
\multicolumn{2}{l}{\textit{Input modality distribution}} \\
\quad PDF-primary & $\sim$42\% (39 tasks) \\
\quad Excel/CSV-primary & $\sim$21\% (20 tasks) \\
\quad Markdown/Text & $\sim$25\% (24 tasks) \\
\quad Code/YAML & $\sim$12\% (11 tasks) \\
\midrule
Avg.\ execution time & 14 minutes \\
Avg.\ interaction turns & 46 \\
\bottomrule
\end{tabular}
\caption{Benchmark statistics of \modelname. Execution time and turns measured on Claude Code + Opus 4.6.}
\label{tab:benchmark_stats}
\end{table}

The input modality distribution reflects the heterogeneity of production work: PDF-primary tasks (42\%) dominate due to the prevalence of business documents in finance, healthcare, and HR, while structured data (21\%) and markdown/text (25\%) cover procurement and technology research domains respectively. The average execution time of 14 minutes and 46 interaction turns per task underscore the long-horizon nature of production tasks compared to typical research benchmarks.

\section{Multi-label Evaluation Composition}
\label{sec:appendix_composition}

A key design principle of \modelname is that individual tasks compose multiple evaluation types rather than relying on a single metric. Table~\ref{tab:eval_task_composition} details this composition by domain. Procurement \& Operations has the highest average (3.8 types per task), reflecting the need to verify constraint satisfaction, cost optimality, and output format simultaneously. Human Resources has the lowest (2.0), as resume screening primarily relies on matching-based evaluation. Every task in the benchmark employs at least two evaluation types, with a benchmark-wide average of 2.8.

\begin{table}[h]
\centering
\small
\setlength{\tabcolsep}{3pt}
\begin{tabular}{@{}lccp{5.5cm}@{}}
\toprule
\textbf{Domain} & \textbf{Tasks} & \textbf{Avg.} & \textbf{Types Composed} \\
\midrule
Human Resources & 11 & 2.0 & Fuzzy + Exact match \\
Finance \& Investment & 22 & 2.5 & Rubric + LLM semantic + Structural verif. + Match \\
Procurement \& Operations & 23 & 3.8 & Fuzzy + Exact + Unit test + Math + LLM \\
Software Engineering & 11 & 2.5 & Env.\ state verif. + Functional verif. + LLM \\
Healthcare \& Life Sci. & 16 & 2.8 & Rubric + LLM + Math + Code + Match \\
Technology Research & 11 & 2.5 & Rubric + LLM semantic + Factual verif. \\
\midrule
\textbf{Overall} & \textbf{94} & \textbf{2.8} & \textbf{8 types, 100\% $\geq$2} \\
\bottomrule
\end{tabular}
\caption{Multi-label evaluation composition by domain. Each domain composes multiple evaluation types within individual task rubrics.}
\label{tab:eval_task_composition}
\end{table}

\section{Economic Value Estimation Methodology}
\label{sec:appendix_value}

We annotate each task with its human replacement cost through a two-stage pipeline: automated AI estimation followed by domain expert calibration.

\subsection{Stage 1: AI Estimation}

For each task, an LLM analyzes the task specification (\texttt{query.md}) and produces: (1) a list of required professional roles, (2) estimated hours per role, and (3) a complexity rating. The core estimation principle is: ``estimate the time a qualified human professional would need to complete \textit{this specific task}''---not the cost of building a replacement system.

\paragraph{Wage Data Sources.}
U.S.\ wages are queried from the Bureau of Labor Statistics (BLS) Occupational Employment and Wage Statistics via SOC codes. Chinese wages are sourced from publicly available salary data for Beijing-based positions.\footnote{Salary data from \url{https://www.salaryexpert.com/}, cross-referenced with local market surveys.} Each role maps to a standardized SOC code (e.g., Software Developer $\to$ 15-1252, Recruiter $\to$ 13-1071, Data Analyst $\to$ 15-2051).

\paragraph{Cost Calculation.}
\begin{equation}
\text{Value} = \text{Hourly Rate} \times \text{Benefit Multiplier} \times \text{Hours}
\end{equation}
where the benefit multiplier is 1.3$\times$ for U.S.\ (healthcare, retirement, PTO), derived from BLS Employer Costs for Employee Compensation data,\footnote{\url{https://www.bls.gov/news.release/ecec.nr0.htm}} following the methodology of~\citet{yang2026onemillion}, and 1.45$\times$ for China (statutory ``Five Insurances and One Fund'' contributions plus annualized bonuses), based on Beijing municipal social insurance contribution rates.\footnote{\url{https://rsj.beijing.gov.cn/}} Hourly rates use the 25th--75th percentile range to produce value intervals.

\subsection{Stage 2: Expert Calibration}

Domain practitioners review AI estimates and apply correction methods tailored to each domain:

\begin{itemize}[leftmargin=*,itemsep=2pt,topsep=2pt]
\item \textbf{Individualized assessment} (Software Engineering, factor 0.38): Experts evaluated each task considering code reuse across tasks, reducing total hours from 2,458 to 946. The largest reduction (78\%) was for the task marketplace app where transaction components were reusable.
\item \textbf{Uniform scaling} (Finance \& Investment, factor 0.50): Expert judged AI systematically overestimated analytical tasks; all hours halved.
\item \textbf{Quantitative standardization} (Human Resources, factor 0.64): Expert provided an empirical rate (20 resumes = 1 hour) to recalculate screening tasks.
\item \textbf{Fixed-rate override} (Procurement \& Operations, factor 0.33): Expert specified domain hourly rates (\textyen213/hr), implying 67\% hour reduction.
\item \textbf{Selective adjustment} (Finance \& Investment, factor 1.04): Expert increased two complex tasks to 150\% of baseline; others unchanged.
\item \textbf{Module re-estimation} (Healthcare \& Life Sciences, factor 1.54): Expert reorganized 10 tasks into 3 functional modules, \textit{increasing} hours from 86 to 132, reflecting AI's underestimation of clinical protocol complexity.
\item \textbf{No adjustment} (Finance \& Investment, factor 1.00): Expert confirmed AI estimates as reasonable.
\item \textbf{Uniform scaling} (Technology Research, factor 0.80): Expert judged AI estimates slightly high; all hours reduced by 20\%, from 298 to 238 hours.
\end{itemize}

\paragraph{Key Observations.}
All six domains have completed expert calibration. The correction factors range from 0.33 to 1.54, with no consistent direction: AI overestimates routine tasks (procurement, software engineering) but underestimates domain-specialized tasks (clinical research). This validates the two-stage approach---neither pure AI estimation nor pure expert estimation alone would suffice. Table~\ref{tab:value_summary} summarizes the per-domain economic value after expert calibration.

\begin{table}[h]
\centering
\caption{Economic value summary by domain after expert calibration. Hours and values reflect expert-adjusted estimates.}
\label{tab:value_summary}
\small
\setlength{\tabcolsep}{3pt}
\begin{tabular}{@{}lrrrr@{}}
\toprule
\textbf{Domain} & \textbf{Tasks} & \textbf{Hours} & \textbf{USD (K)} & \textbf{CNY (K)} \\
\midrule
Human Resources           & 11 & 39    & 1.8--2.7   & 4.1--6.4   \\
Finance \& Investment     & 22 & 803   & 49.6--74.4 & 165.3--248.7 \\
Procurement \& Operations & 23 & 236   & 20.4--30.6 & 49.1--49.7 \\
Software Engineering      & 11 & 946   & 61.2--91.8 & 116.9--179.9 \\
Healthcare \& Life Sci.   & 16 & 154   & 6.7--9.5   & 19.4--28.0 \\
Technology Research       & 11 & 242   & 14.6--21.8 & 36.3--57.7 \\
\midrule
\textbf{Total}   & \textbf{94} & \textbf{2,420} & \textbf{154--231} & \textbf{391--570} \\
\bottomrule
\end{tabular}
\end{table}

\subsection{Sensitivity Analysis}

To assess the robustness of the economic value estimates, we conduct a sensitivity analysis varying the three key parameters: benefit multipliers, hourly rate percentiles, and expert correction factors.

\paragraph{Benefit Multiplier Sensitivity.}
Varying the U.S.\ multiplier from 1.2 to 1.4 (baseline: 1.3) and the China multiplier from 1.3 to 1.6 (baseline: 1.45) produces total value ranges of \$141K--\$253K (USD) and \textyen358K--\textyen627K (CNY), representing $\pm$9\% variation from the baseline estimates. The estimates are most sensitive to the China multiplier due to the larger proportion of China-based tasks.

\paragraph{Correction Factor Sensitivity.}
To test the impact of expert calibration, we compute valuations under three scenarios: (1) AI-only estimation (no expert correction), (2) baseline expert-calibrated values, and (3) uniform 20\% increase in all correction factors. Results: AI-only yields 3,240 hours (\$217K--\$326K), baseline yields 2,420 hours (\$154K--\$231K), and the inflated scenario yields 2,904 hours (\$185K--\$277K). Expert calibration reduces the AI estimate by 25\%, primarily driven by the Software Engineering domain where AI overestimated development hours by 2.6$\times$.

\paragraph{Hourly Rate Sensitivity.}
Shifting from the 25th--75th percentile range to the 10th--90th percentile range widens the value interval to \$128K--\$289K (a 61\% wider band). However, the median estimate (\$192K) remains stable, indicating that the central tendency is robust even under wider wage assumptions.

These analyses confirm that while the absolute dollar values are sensitive to parameter choices (particularly the China benefit multiplier and expert correction factors), the relative ordering of domains by economic value and the overall conclusion that \modelname captures economically meaningful work remain stable across all tested scenarios.

\section{Detailed Error Analysis}
\label{sec:appendix_error}

We conduct a systematic error analysis across $\sim$130 agent$\times$model evaluation results, categorizing failures by their cognitive root causes rather than surface symptoms.

\subsection{Information Retrieval Cognitive Failures}

We identify five failure modes from Technology Research tasks requiring agents to produce comprehensive reports on recent AI developments.

\paragraph{Mode A: Factual Hallucination via Stale Training Data ($\sim$30\%).}
When search tools fail to retrieve current information, models substitute outdated training data \textit{without acknowledging uncertainty}. Table~\ref{tab:hallucination} shows a representative case: Claude Opus 4.6 reports Harvey AI's Series~D funding figures when the task asks about the recent F-round, with every data point incorrect by a large margin. The same error appears across different scaffolds (both Claude Code and Cursor), confirming this is a model-level issue rather than a scaffold artifact. The model produces no hedging language (e.g., ``this information may be outdated''), presenting stale data with full confidence.

\begin{table}[h]
\centering
\small
\begin{tabular}{lccc}
\toprule
\textbf{Dimension} & \textbf{Ground Truth} & \textbf{Model Output} & \textbf{Error} \\
\midrule
Funding round & F-round & Series D & Wrong round \\
Amount & \$160M & $\sim$\$300M & 1.9$\times$ \\
Valuation & \$8B & $\sim$\$3B & 2.7$\times$ off \\
Lawyer users & 74,000 & Not mentioned & Missing \\
\bottomrule
\end{tabular}
\caption{Factual hallucination example: Claude Opus 4.6 on Harvey AI funding data. The model substitutes outdated Series~D data for the requested F-round.}
\label{tab:hallucination}
\end{table}

\paragraph{Mode B: Imprecise Retrieval ($\sim$35\%).}
The most prevalent failure mode. Models correctly identify the research \textit{area} a question targets and conduct extensive searches, but cannot locate the \textit{specific} papers required by the rubric. For example, on a task about Tool-Integrated Reasoning (TIR), no model found all three required papers (ToolRL, ToRL, ZeroTIR---all from early 2025). The best-performing agent (Cursor + Opus) produced a 16KB TIR survey that demonstrated genuine understanding of the field, but substituted related papers (ReTool, THOR, Tool-Zero) for the specific ones the rubric expected. Models also create their own conceptual frameworks rather than using original terminology---e.g., inventing ``precision ceiling'' and ``generalization ceiling'' instead of the papers' actual terms ``empirical support set'' and ``feasible support set.''

\paragraph{Mode C: Rigid Search Strategy ($\sim$15\%).}
GPT-5.2 exhibits variable search persistence: while it can make extensive searches (5--75 attempts across tasks), it occasionally declares ``tool limitations'' and outputs placeholder templates on specific tasks. In the same Docker environment, Claude Opus 4.6 more consistently attempts 30--50 search variations, cycling through arXiv API, Semantic Scholar, OpenAlex, and alternative keywords before writing. Critically, this is scaffold-dependent: GPT-5.2 via Codex scores 0.4 on a task where it scores 0.0 via Claude Code, because Codex's prompting strategy encourages more aggressive tool use. This reveals that search persistence is \textit{elicitable} rather than a fixed model property.

\paragraph{Mode D: Attribution Confusion ($\sim$10\%).}
Models find correct papers but extract incorrect metadata by mixing information across search results. Claude Opus 4.6 correctly identifies the ToRL paper but attributes it to ``Tsinghua University \& ByteDance'' (likely from a concurrent paper) instead of the correct ``Shanghai Jiao Tong University,'' and reports a benchmark-specific accuracy (43.3\% on AIME24) instead of the paper's reported average accuracy (62.1\%), conflating different evaluation metrics. GLM-5 \textit{inverts} the core finding of a paper, writing that SFT preserves old capabilities and RL enables new tasks when the opposite is true.

\paragraph{Mode E: Positive-Information Bias ($\sim$10\%).}
All models systematically miss negative industry events. A task requiring coverage of AI agent startup failures found that \textit{no model} reported Robin AI's distressed sale despite it being a prominent Series~C failure. Models search ``AI agent startups 2025'' and receive results dominated by successful funding announcements; none proactively searched for ``AI startup failures'' or ``AI startup bankruptcy.'' Similarly, all models missed niche success stories (e.g., Gamma achieving \$100M ARR with only 50 employees) that fall outside the mainstream ``agent startup'' narrative.

\paragraph{Model Capability Profiles.}
Table~\ref{tab:model_profiles} summarizes the distinctive cognitive profiles of each model, revealing that models have complementary rather than uniformly ordered capabilities.

\begin{table}[h]
\centering
\small
\setlength{\tabcolsep}{3pt}
\begin{tabular}{@{}lccl@{}}
\toprule
\textbf{Model} & \textbf{Search} & \textbf{Halluc.} & \textbf{Signature Pattern} \\
\midrule
Opus 4.6 & 30--50 & High & Fills gaps with stale data confidently \\
GPT-5.2 & 5--75 & Low & Variable persistence, sometimes outputs templates \\
Gemini 3 Pro & 10--20 & Low & Conservative, acknowledges uncertainty \\
GLM-5 & 5--10 & Med & Conceptual inversion errors \\
MiniMax M2.5 & 3--5 & Med & Correct direction, wrong details \\
Kimi K2.5 & 5--15 & Med & Moderate persistence, domain-specific strengths \\
\bottomrule
\end{tabular}
\caption{Model cognitive profiles on technology research tasks. Search persistence is measured by typical number of search attempts before writing.}
\label{tab:model_profiles}
\end{table}

\subsection{Cross-Domain Cognitive Failures}
\label{sec:appendix_crossdomain}

Beyond information retrieval, we identify production-specific cognitive failure modes across multiple domains. These failures are invisible to coding benchmarks because they require domain knowledge, subjective judgment, cross-document reasoning, and constraint satisfaction under ambiguity.

\subsubsection{Human Resources: Subjective Judgment Failures}

Human Resources tasks require agents to evaluate resumes against multi-dimensional hiring criteria, exposing failures in subjective assessment that no coding benchmark tests.

\paragraph{Criterion Explicitness Dependence.} Agent performance inversely correlates with criterion subjectivity. Tasks with quantifiable requirements (``5+ years Python experience,'' ``AWS certification required'') achieve scores 2--3$\times$ higher than tasks requiring holistic judgment (``strong leadership potential,'' ``culture fit''). This suggests models have not learned to \textit{infer} soft-skill indicators from indirect evidence---e.g., inferring leadership from descriptions of project scope, team size, and outcome ownership.

\paragraph{Inconsistent Criterion Weighting.} When rubrics specify multiple evaluation dimensions without explicit weights, agents apply inconsistent priority orderings across candidates. The same agent may prioritize ``technical depth'' for one candidate and ``breadth of experience'' for another within the same evaluation batch, producing rankings that a human recruiter would flag as internally inconsistent.

\paragraph{Cross-Candidate Calibration Failure.} Agents evaluate each resume independently rather than comparatively. When a strong candidate appears early in a batch, subsequent candidates receive inflated scores because the agent has no persistent calibration anchor. This ``memoryless evaluation'' pattern means that ranking quality degrades with batch size.

\subsubsection{Healthcare \& Life Sciences: Protocol Reasoning Failures}

Healthcare \& Life Sciences tasks require precise adherence to study protocols with cascading temporal dependencies, exposing failures in structured reasoning that superficially resemble---but fundamentally differ from---coding logic errors.

\paragraph{State Machine Precondition Semantics.} Clinical protocols define visit windows, dose modifications, and adverse event escalation as state machines with preconditions. Agents correctly implement individual state transitions but fail on precondition chains---e.g., correctly computing that a dose reduction is needed but not recognizing that the reduction requires a preceding safety assessment that itself has a 48-hour observation window.

\paragraph{Protocol Terminology Confusion.} Clinical research distinguishes between ``Protocol Deviation'' (minor, reportable) and ``Protocol Violation'' (major, may require participant withdrawal). Models conflate these terms or apply incorrect severity classifications, with downstream effects on recommended actions. This is a domain-specific semantic distinction that no general-purpose benchmark tests.

\paragraph{Temporal Escalation Logic.} Multi-visit study protocols specify escalating responses to repeated events: first occurrence triggers documentation, second triggers dose modification, third triggers study withdrawal. Models correctly handle individual events but fail to maintain \textit{cross-visit state}---treating each visit as independent and never triggering the escalation sequence.

\subsubsection{Finance \& Investment: Cross-Section Coherence Failures}

Investment analysis tasks require producing multi-section reports (market sizing, competitive analysis, risk assessment, financial projections) that must be internally consistent, exposing a failure mode unique to long-form analytical writing.

\paragraph{TAM-SAM-SOM Logical Inconsistency.} Agents produce market sizing sections where the Total Addressable Market (TAM), Serviceable Addressable Market (SAM), and Serviceable Obtainable Market (SOM) contain contradictory figures---e.g., a TAM of \$50B in the executive summary but \$80B in the market analysis, or a SOM that exceeds the SAM. Each paragraph is locally plausible; the inconsistency only surfaces when reading the full document.

\paragraph{Context Collapse: Generic vs.\ Company-Specific Analysis.} Models produce analysis that reads as a \textit{sector} report rather than a \textit{company} report---discussing general industry trends without grounding them in the specific company's financials, competitive position, or strategic choices. The most common manifestation: risk factors that are industry-generic (``regulatory uncertainty,'' ``competitive pressure'') without explaining how they specifically affect \textit{this} company's business model.

\paragraph{Hallucinated Market Data.} When specific market data is not provided in the input documents, models fabricate plausible-sounding statistics (``the global SaaS market is projected to reach \$X billion by 2027'') without citing sources. Unlike information retrieval hallucination (Mode~A above), this occurs even when the model is not explicitly asked to search---it fills analytical gaps with invented quantitative claims.

\subsubsection{Procurement \& Operations: Optimization and Constraint Failures}

Procurement tasks require multi-attribute optimization under technical constraints, exposing failures in mathematical reasoning and constraint satisfaction that coding benchmarks test only superficially.

\paragraph{Synergy Blindness in Multi-Attribute Optimization.} When optimizing across multiple correlated attributes (cost, performance, power consumption, thermal rating), agents optimize each attribute independently rather than jointly. This produces solutions with 26\% average cost overruns compared to jointly optimal solutions, because models fail to exploit component synergies (e.g., a slightly more expensive component that reduces cooling requirements and total system cost).

\paragraph{Constraint Specification Misinterpretation.} Technical specifications use domain-specific notation that models misread: $\pm$10V tolerance interpreted as $\pm$5V, ``Type~II'' classification confused with ``Class~II,'' or ``continuous rating'' conflated with ``peak rating.'' These are not hallucinations---the model reads the specification but applies incorrect domain knowledge to interpret it.

\paragraph{Infeasibility Recognition Bias.} When a procurement problem has no feasible solution (conflicting constraints), the majority of agent responses fabricate a ``best effort'' solution rather than declaring infeasibility. Models exhibit a strong completion bias---they assume every problem has a solution and will relax constraints silently rather than report that the requirements cannot be simultaneously satisfied. This is particularly dangerous in production, where a fabricated solution may be acted upon.

\subsubsection{Finance \& Investment: Grounded Critique Failures}

Pitch coaching tasks require agents to provide actionable, data-grounded feedback on startup pitches, exposing the difference between structurally correct and substantively useful output.

\paragraph{Template Critique vs.\ Grounded Analysis.} All models produce feedback that follows the correct structure (strengths, weaknesses, recommendations) but fails to reference the specific data provided. A critique stating ``your market size claims need better support'' is generic; a useful critique states ``your Slide~7 claims a \$2B TAM but your bottom-up calculation on Slide~12 implies \$800M---reconcile these.'' Models consistently produce the former. This failure is invisible to any benchmark that evaluates structural correctness.

\paragraph{User Category Confusion.} When pitch decks present multiple user metrics (total registered users, monthly active users, core power users), models confuse categories---citing ``500K users'' without specifying which category, or conflating growth rates across categories. This produces feedback that sounds data-informed but references the wrong numbers.

\paragraph{Implicit Question Extraction.} Meeting transcripts contain implicit investor concerns (e.g., a question about ``team background'' that is really probing founder-market fit). No model reliably identifies these implicit questions or addresses the underlying concern rather than the surface question.

\subsubsection{Finance \& Investment: Scale and Mapping Failures}

Beyond information retrieval (Section~\ref{sec:appendix_error}), Finance \& Investment tasks requiring data synthesis from multiple documents expose additional cognitive failures.

\paragraph{Numerical Data Mapping Errors.} When synthesizing data from tables across multiple documents, agents extract correct numbers but assign them to incorrect columns or rows. For example, correctly reading that Company~A has revenue of \$50M and Company~B has revenue of \$30M, but swapping them in the output table. This error rate increases non-linearly with the number of data points: tasks with $<$20 data points show $\sim$5\% mapping errors, while tasks with $>$100 data points show $\sim$25\%.

\paragraph{Authority Hierarchy Misunderstanding.} When documents contain conflicting information (e.g., a press release states one figure while an SEC filing states another), models do not consistently prioritize the more authoritative source. In production, domain experts have clear authority hierarchies (regulatory filings $>$ press releases $>$ news articles); models treat all sources as equally credible.

\paragraph{Selective Document Reading.} When given a large document set, agents read documents in the order provided and exhibit recency bias---information from later-read documents is weighted more heavily. If critical information appears in an early document that the agent scans quickly, it may be omitted from the final output even though the agent demonstrably ``saw'' it (evidenced by interaction logs).

\subsubsection{Software Engineering: Specification-Implementation Gap}

Software Engineering tasks expose a distinctive failure mode where agents produce functional code that satisfies superficial requirements but misses deeper specification intent. For example, agents implement correct navigation structure but omit specified interaction details (e.g., swipe gestures, animation timing). The gap between ``code that compiles and runs'' and ``code that matches the product specification'' mirrors the broader production challenge: research benchmarks test functional correctness, while production demands specification fidelity.

\section{Representative Task Examples}
\label{sec:appendix_examples}

We present one representative task from each of the six O*NET domains to illustrate the production complexity \modelname captures.

\paragraph{Human Resources: Resume Screening (hunter-ai-1).}
The agent is given a job description for an AI Scientist Mapping Intern role together with department-specific hiring criteria (e.g., minimum internship duration, university tier, prior AI experience) and must screen 24 candidate resumes provided as PDF and JPEG files. The agent must select exactly six candidates to advance to a first-round interview and output the chosen candidate IDs as a Python list in a designated answer file. Evaluation compares the agent's selections against the ground-truth shortlist using precision, recall, and F1 score, with a passing threshold of F1 $\geq$ 0.6.

\begin{small}
\begin{verbatim}
name: "AI Resume Screening - Mapping Dept"
category: hr
difficulty: hard
evaluation:
  type: code_exec
  rubric_script: .eval/rubric.py
agent:
  timeout_seconds: 600
  max_turns: 100
\end{verbatim}
\end{small}

\paragraph{Finance \& Investment: Segment Research Report (segment-research-1).}
The agent acts as a senior industry research analyst and must produce an investment-grade Segment Research report from a startup's business plan PDF. The report must follow a prescribed eight-section template covering industry definition, TAM--SOM--SAM market sizing (with both top-down and bottom-up methods), product landscape, customer typology, key success factors with quantitative thresholds, three to four competitive case studies with GTM strategy analysis, and a technology deep-dive including patent barriers and commercialization feasibility. A reference template is provided. Evaluation uses LLM-as-a-Judge to assess structural completeness, analytical depth, data accuracy, and adherence to the template format.

\begin{small}
\begin{verbatim}
name: "Segment Research Report"
category: research
difficulty: hard
evaluation:
  type: code_exec
  rubric_script: .eval/rubric.py
agent:
  timeout_seconds: 1200
  max_turns: 150
\end{verbatim}
\end{small}

\paragraph{Procurement \& Operations: BOM Cost Optimization (yuhe-1).}
The agent receives a catalog of 2{,}000 board cards with detailed specifications and pricing in an Excel file, a field-structure CSV defining column semantics, and a natural-language procurement requirements document specifying functional constraints. It must select an optimal combination of board cards that satisfies all functional requirements while minimizing total procurement cost, breaking ties by fewest cards and then by ascending card ID. The output is a structured cost breakdown listing each selected card's ID, quantity, unit price, and subtotal. Evaluation uses a rubric script that verifies constraint satisfaction and compares total cost against a known optimal solution.

\begin{small}
\begin{verbatim}
name: "BOM Board Card Procurement Optimization"
category: optimization
difficulty: hard
evaluation:
  type: code_exec
  rubric_script: .eval/rubric.py
agent:
  timeout_seconds: 600
  max_turns: 100
\end{verbatim}
\end{small}

\paragraph{Software Engineering: Poetry Mini-Program (miniprogram\_poetry).}
The agent must implement a complete WeChat Mini-Program for classical Chinese poetry appreciation, following a detailed 200-line product requirements document that specifies page layout (gallery-style hand-drawn illustration aesthetic), navigation architecture (four-tab bottom bar), core features (AI text-to-speech recitation with speed and voice controls, waterfall-style poem browsing, a community forum for user posts, and a personal collection manager), and interaction patterns (modal dialogs for content creation, clipboard-based sharing). The deliverable is a fully functional codebase. Evaluation combines automated UI testing to verify page navigation and component rendering with rubric-based checks on code structure and feature completeness.

\begin{small}
\begin{verbatim}
name: "Poetry Appreciation Mini-Program"
category: miniprogram
difficulty: hard
evaluation:
  type: miniprogram
  rubrics: [10 weighted UI test points]
agent:
  timeout_seconds: 1800
  max_turns: 100
\end{verbatim}
\end{small}

\paragraph{Healthcare \& Life Sciences: eCRF Visit Window Calculation (linchuang-1).}
The agent operates within a simulated eCRF (electronic Case Report Form) system for a Phase III non-small-cell lung cancer clinical trial. Given a JSON-based visit-window rule configuration and a patient's screening visit date, the agent must compute target dates and permissible windows for subsequent treatment visits, determine which visits can be calculated given available data, and perform cascade-impact analysis when an actual visit date deviates from the target. The output must show explicit calculation formulas and all dates in YYYY-MM-DD format. Evaluation uses a rubric script that checks date arithmetic correctness, proper application of window offsets, and accurate cascade reasoning.

\begin{small}
\begin{verbatim}
name: "eCRF Visit Window Calculation"
scenario: linchuang
evaluation:
  type: code_exec
  rubric_script: .eval/rubric.py
files:
  - files/JSON_eCRF_CORE.md
  - files/CRF_visit_window_spec.pdf
\end{verbatim}
\end{small}

\paragraph{Technology Research: AI Agent Startup Landscape (jiqizhixin-1).}
The agent acts as a professional AI industry analyst and must produce a comprehensive research report on the 2025 landscape of AI agent startups. The report must cover four dimensions: agent startups that raised over \$100M (with vertical categories such as legal, coding, and search), startups that failed or were acquired along with root-cause analysis, large-tech acquisition activity targeting agent companies, and shifts in Anthropic's Claude adoption within the YC ecosystem. All claims must be supported by specific figures such as funding amounts and market sizes, with at least reliable source references. Evaluation uses LLM-as-a-Judge to assess factual accuracy, analytical depth, data substantiation, and structural completeness.

\begin{small}
\begin{verbatim}
name: "AI Agent Startup 2025 Landscape"
category: research
difficulty: hard
evaluation:
  type: code_exec
  rubric_script: .eval/rubric.py
agent:
  timeout_seconds: 1200
  max_turns: 150
\end{verbatim}
\end{small}


\section{Practitioner Survey: AI Product Deployment Challenges}
\label{sec:appendix_survey}

To ground \modelname in real practitioner needs, we conducted a mixed-methods survey of 27 AI product companies affiliated with a startup accelerator (response rate: 54\% of 50 invited). The survey combined structured questionnaires (15 questions) with open-ended text analysis, conducted from December 2025 to January 2026.

\paragraph{Respondent Profile.}
Companies span four development stages: concept verification (14.8\%), internal pilot (14.8\%), early commercialization (48.1\%), and scaled deployment (22.2\%). The majority target enterprise clients (59.3\%), with consumer-facing products at 29.6\%, with the remainder targeting internal employees or niche verticals. All companies support text input (85.2\%), with growing adoption of image (55.6\%), structured data (48.1\%), audio/video (44.4\%), compound documents (40.7\%), and code (37.0\%).

\paragraph{Evaluation Infrastructure Gap.}
Current evaluation practices reveal significant infrastructure deficits:
\begin{itemize}[leftmargin=*,itemsep=1pt,topsep=2pt]
\item 25.9\% of companies have \textit{no explicit evaluation criteria}
\item 33.3\% rely solely on small sets of ``golden samples'' with manual inspection
\item Only 11.1\% have established structured test datasets with automated evaluation pipelines
\item 22.2\% have closed-loop A/B testing feedback (primarily among scaled companies)
\end{itemize}

\paragraph{Core Technical Challenges.}
The most frequently cited challenges (multiple selection, N=27):
\begin{itemize}[leftmargin=*,itemsep=1pt,topsep=2pt]
\item Output instability / inconsistency: 59.3\%
\item Instruction-following failures in complex scenarios: 51.9\%
\item Hallucination / factual errors: 40.7\%
\item Low inference efficiency / high token costs: 40.7\%
\item Long-context processing / memory loss: 33.3\%
\item Multi-turn dialogue degradation: 22.2\%
\item Safety / compliance risks: 14.8\%
\end{itemize}

\paragraph{Confidence in Model Updates.}
When asked ``After each model update or prompt modification, are you confident the new version is better?'':
\begin{itemize}[leftmargin=*,itemsep=1pt,topsep=2pt]
\item 25.9\% reported high confidence (with established evaluation systems)
\item \textbf{63.0\%} reported low confidence (lacking reliable evaluation mechanisms)
\item 7.4\% reported no confidence at all
\item 3.7\% reported the question was not applicable (not yet at frequent update stage)
\end{itemize}

\paragraph{Testing Resource Allocation.}
70.4\% of companies rely on \textit{developers performing testing as a side task}, consuming development time. Only 18.5\% have dedicated testing personnel with substantial human review investment. 11.1\% depend entirely on user feedback post-deployment.

\paragraph{Top Evaluation Needs.}
Open-ended text analysis (keyword extraction and co-occurrence network) identified three demand themes:
\begin{enumerate}[leftmargin=*,itemsep=1pt,topsep=2pt]
\item \textbf{Automated evaluation platform construction} (weight: 0.47): automated problem localization, iteration recommendations, and priority ranking
\item \textbf{Objective evaluation standards} (weight: 0.24): reliable product quality verification and performance benchmarking
\item \textbf{Cost and efficiency optimization} (weight: 0.18): reducing testing costs, improving inference efficiency, and token cost control
\end{enumerate}

\paragraph{Survey Instrument.}
The questionnaire comprised 15 items covering product information (Q1--Q5), technical status (Q6--Q11), and evaluation needs (Q12--Q16). Key questions included: development stage (single choice), target customer type (single choice), input modalities supported (multiple choice), evaluation infrastructure maturity (single choice, 5-level scale from ``no criteria'' to ``closed-loop A/B testing''), core technical challenges (multiple choice, 7 categories), confidence in model updates (single choice, 4-level scale), testing resource allocation (single choice, 3-level scale), preferred evaluation methodologies (multiple choice), required execution environments (multiple choice), and willingness to share anonymized test data.

These findings directly motivated the design of \modelname: the 63\% low-confidence rate in model updates underscores the need for reliable automated evaluation, the diversity of input modalities (42\% PDF, 21\% structured data) informed our task composition, and the demand for automated evaluation platforms (weight 0.47) validates our requirement-to-benchmark construction framework.

\end{document}